\documentclass[lettersize,journal]{IEEEtran}
\usepackage{cite}
\usepackage{amsmath,amssymb,amsfonts}
\usepackage{algorithmic}
\usepackage{algorithm}
\usepackage{graphicx}
\usepackage{textcomp}
\usepackage{xcolor}
\usepackage{soul}
\usepackage{hyperref}
\usepackage{fancyhdr}
\usepackage{float}
\usepackage{booktabs}
\usepackage{multirow}
\usepackage{makecell}
\usepackage{caption}
\usepackage[caption=false,font=normalsize,labelfont=sf,textfont=sf]{subfig}
\usepackage{subcaption}
\usepackage[section]{placeins}
\usepackage{multicol}

\begin{document}
\title{D-FaST: Cognitive Signal Decoding \\ with Disentangled Frequency-Spatial-Temporal Attention
}

\author{WeiGuo Chen, Changjian Wang, Kele Xu, Yuan Yuan, Yanru Bai and Dongsong Zhang
    \thanks{WeiGuo Chen, Changjian Wang, Kele Xu, Yuan Yuan are with National University of Defense Technology, Changsha, Hunan, 410000, China. Email: chenweiguo@nudt.edu.cn, wangcj@nudt.edu.cn, xukelele@163.com, yuanyuan@nudt.edu.cn}
    \thanks{Yanru Bai is with Academy of Medical Engineering and Translational Medicine, Tianjin University, Tianjin, 300072, China. Email: yr56\_bai@tju.edu.cn}
    \thanks{Dongsong Zhang is with School of Big Data and Artificial Intelligence, Xinyang College, Xinyang, Henan, 464000, China. Email: dszhang@nudt.edu.cn}
    \thanks{Corresponding author: Changjian Wang, Kele Xu}
    \thanks{Code is available at \href{https://github.com/AdFiFi/D-FaST.git}{https://github.com/AdFiFi/D-FaST.git}}
}

\markboth{Weiguo \MakeLowercase{\textit{et al.}}: D-FaST: Cognitive Signal Decoding with Disentangled Frequency-Spatial-Temporal Attention}
{Weiguo \MakeLowercase{\textit{et al.}}: D-FaST: Cognitive Signal Decoding with Disentangled Frequency-Spatial-Temporal Attention}

\maketitle
\begin{abstract}
    Cognitive Language Processing (CLP), situated at the intersection of Natural Language Processing (NLP) and cognitive science, plays a progressively pivotal role in the domains of artificial intelligence, cognitive intelligence, and brain science. Among the essential areas of investigation in CLP, Cognitive Signal Decoding (CSD) has made remarkable achievements, yet there still exist challenges related to insufficient global dynamic representation capability and deficiencies in multi-domain feature integration. In this paper, we introduce a novel paradigm for CLP referred to as Disentangled Frequency-Spatial-Temporal Attention(D-FaST). Specifically, we present an novel cognitive signal decoder that operates on disentangled frequency-space-time domain attention. This decoder encompasses three key components: frequency domain feature extraction employing multi-view attention, spatial domain feature extraction utilizing dynamic brain connection graph attention, and temporal feature extraction relying on local time sliding window attention. These components are integrated within a novel disentangled framework. Additionally, to encourage advancements in this field, we have created a new CLP dataset, MNRED. Subsequently, we conducted an extensive series of experiments, evaluating D-FaST's performance on MNRED, as well as on publicly available datasets including ZuCo, BCIC IV-2A, and BCIC IV-2B. Our experimental results demonstrate that D-FaST outperforms existing methods significantly on both our datasets and traditional CSD datasets including establishing a state-of-the-art accuracy score 78.72\% on MNRED, pushing the accuracy score on ZuCo to 78.35\%, accuracy score on BCIC IV-2A to 74.85\% and accuracy score on BCIC IV-2B to 76.81\%.
\end{abstract}


\section{Introduction}

\begin{figure}[!t]
    \centering
    \includegraphics[width=3in]{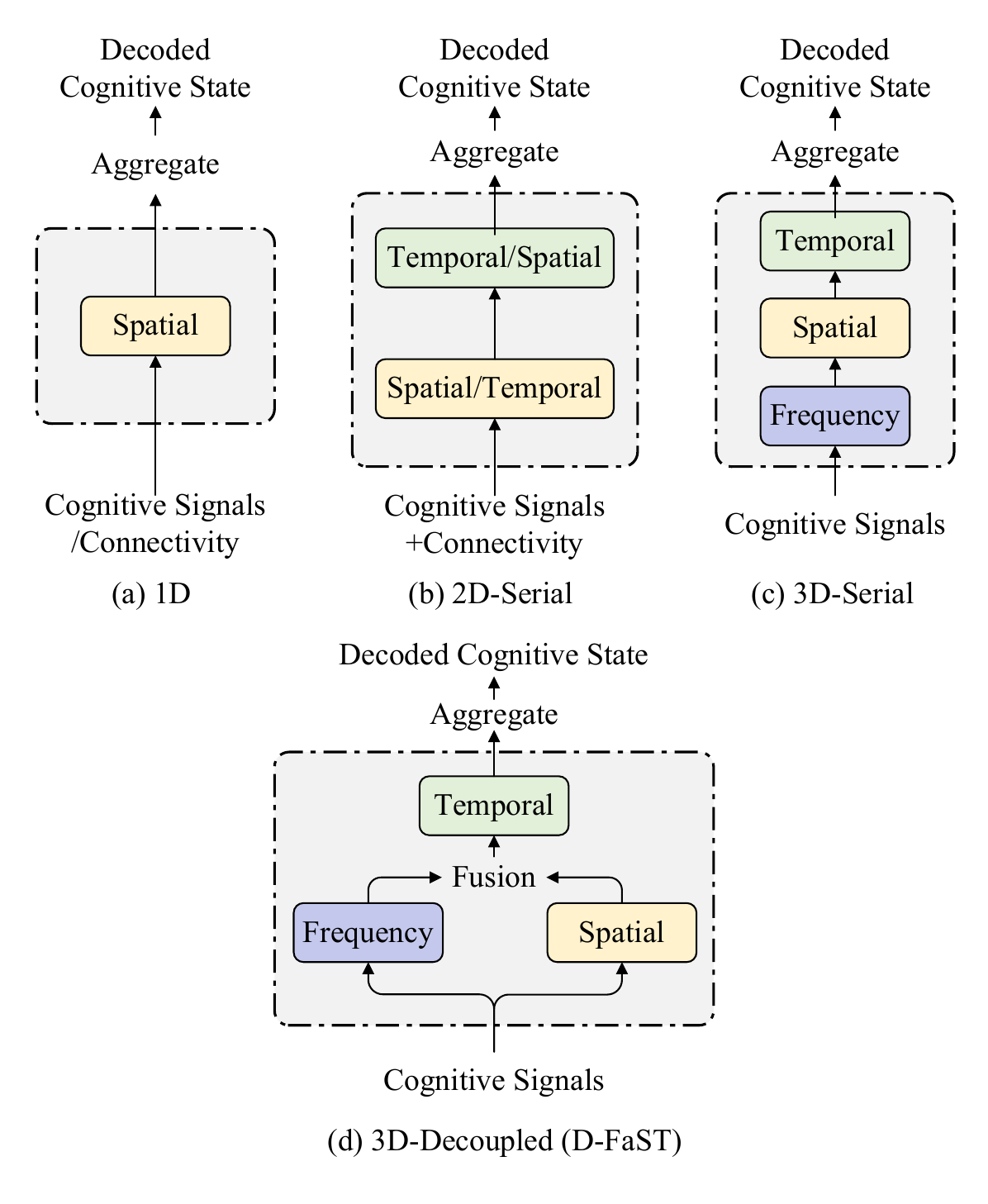}
    \caption{Conceptual comparison of four brain signal decoding architectures. (a): The Single-Domain (1D) Architecture primarily focuses on the extraction of spatial domain information from cognitive signals. (b): The Double-Domain (2D) Serial Architecture predominantly extracts both spatial and temporal domain information, either in different orders or simultaneously. (c): The Triple-Domain (3D) Serial Architecture sequentially extracts information from the frequency domain, spatial domain, and temporal domain. (d): The Triple-Domain Disentangled Architecture initially processes cognitive signals through the frequency and spatial domains, resulting in separate frequency and spatial features.}
    \label{fig_3D}
\end{figure}

\IEEEPARstart{C}{ognitive} Signal Decoding (CSD), a fundamental domain within Cognitive Language Processing (CLP), assumes a pivotal role in the context of few-shot learning~\cite{R64}, interpretable deep learning-based Natural Language Processing (NLP)~\cite{R63, R65, R66}, and delving into the intricacies of language physiology in the human brain, thus contributing to the field of neuro-prosthesis~\cite{R67, R68}. CSD, particularly when coupled with neuro-imaging techniques such as Electroencephalography (EEG) and functional Magnetic Resonance Imaging (fMRI), has emerged as an indispensable tool for researchers delving into cognitive science. Among the neuro-imaging modalities, EEG stands out as one of the most commonly employed methods in CLP due to its high temporal resolution. Consequently, several deep learning techniques have surfaced as the primary means of CSD, leading to substantial progress in this domain~\cite{R1, R30, R41, R87}.

\begin{figure}[!t]
    \centering
    \includegraphics[width=\linewidth]{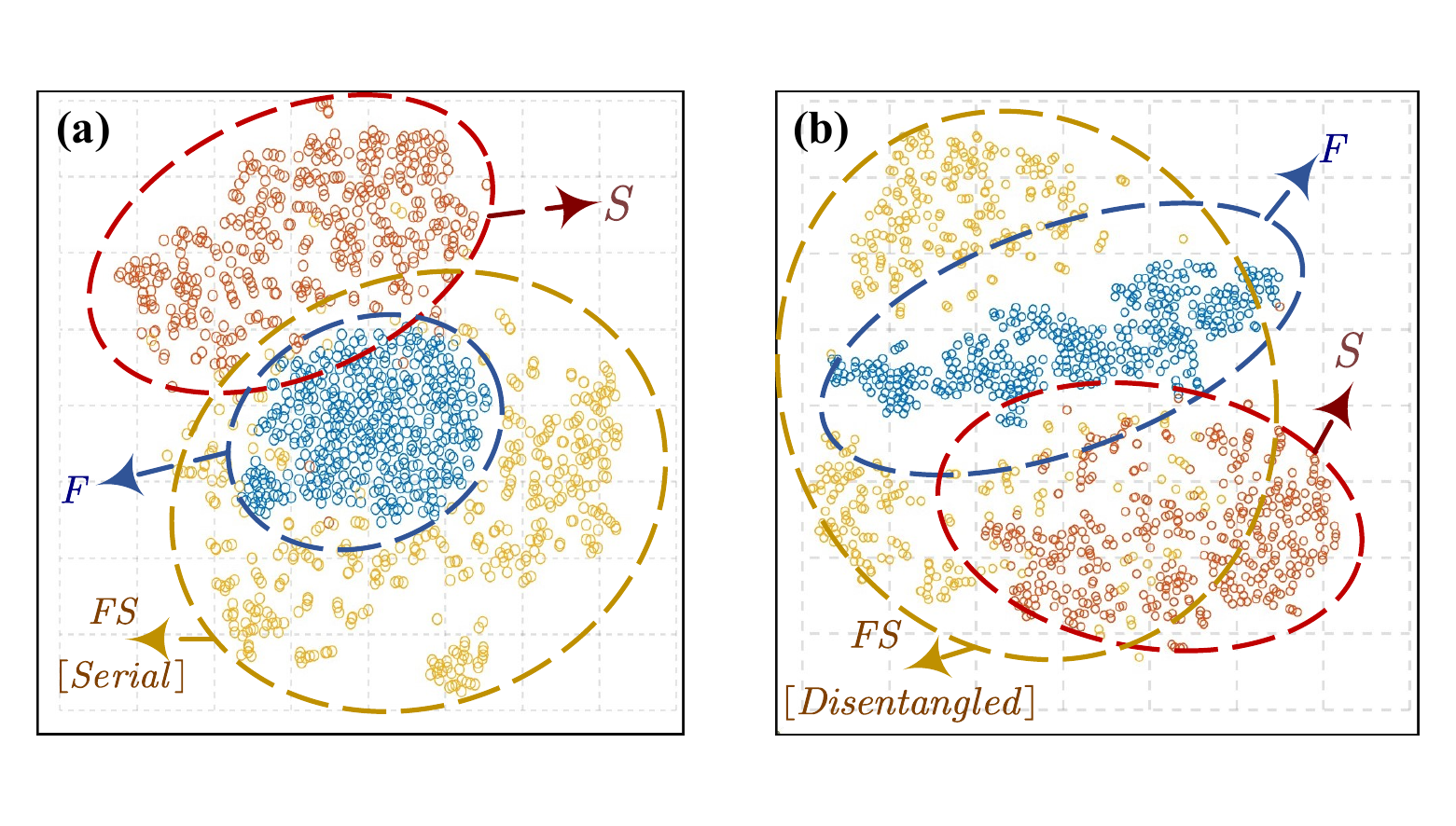}
    \caption{t-SNE projections of feature extracted by EEGNet~\cite{R1} with different strategies: (a) Serial Framework(Vanilla), (b) Disentangled Framework(Ours). The dashed circles indicate the range of projected features. The visualization details can be found in our open source code.}
    \label{fig_FS_TSNE}
\end{figure}

EEG signals exhibit intricate characteristics across frequency, spatial, and temporal domains, particularly in the context of CLP. The question of how to effectively extract features from these multiple domains and construct mechanisms for their integration require thorough examination. Currently, three primary frameworks, as depicted in Fig.~\ref{fig_3D}, have been established based on the incorporation of information from different domains and fusion methods. The first framework, referred to as the single-domain (1D) architecture~\cite{R3, R11, R43}, places a significant emphasis on the connectivity of cognitive signals, predominantly extracting information from the spatial domain. The second framework, known as the double-domain serial (2D) architecture~\cite{R34, R30, R24, R4, R48, R41, R42}, primarily extracts information from both the spatial and temporal domains in varying orders~\cite{R30, R24, R4, R48, R41}, or simultaneously~\cite{R34, R42}. The third framework, the triple-domain serial (3D) architecture, sequentially extracts information from the frequency domain, spatial domain, and temporal domain~\cite{R1}.
However, it is noteworthy that when it comes to extracting information from both the frequency and spatial domains, or from both the temporal and spatial domains, most models opt for a sequential approach~\cite{R1, R4, R5, R34, R48, R42, R23}. These methods may neglect the observation that the spatial domain shares less relevance but greater independence with the other two domains, as they are orthogonal in dimension. In contrast, the frequency domain shares less independence but greater relevance with the temporal domain, as they offer distinct perspectives on time series information. Consequently, the sequential extraction of features from different domains may disrupt the overall extraction process. Fig.~\ref{fig_FS_TSNE} intuitively presents the feature distributions of EEGNet~\cite{R1} under the original serial framework and the disentangled framework. The frequency-spatial features obtained by the vanilla EEGNet evidently fail to adequately represent the task-specific frequency and spatial characteristics inherent in the data. In contrast, the disentangleded approach effectively encapsulates these aspects.

Convolutional Neural Networks (CNNs) have demonstrated notable advantages in extracting intricate information. Several widely recognized CNN-based models for Cognitive Signal Decoding (CSD)~\cite{R1, R2, R3, R4, R5, R6, R34, R43, R48, R41, R42} are dedicated to enhancing CNNs' performance in the context of CSD. However, human brain cognitive processes exhibit substantial contextual relevance and generally have longer duration compared to other processes, such as Event-Related Potential (ERP) or Error-Related Negativity potentials (ERN). Simultaneously, for the sake of facilitating matrix operations, it is customary to represent signals collected by sensors positioned in three-dimensional space using two-dimensional multivariate time series.
On one hand, convolution operations, renowned for their local feature extraction capabilities, encounter difficulties in capturing disrupted adjacency relationships between nodes. On the other hand, even when nodes are physically adjacent, convolution operations struggle to effectively capture functional connections between non-adjacent nodes. Many researchers have sought to enhance cognitive signal decoders by incorporating Transformers~\cite{R64, R73, R70, R71, R11, R18}, recognizing their proficiency in representing global and contextual features, and their remarkable progress in NLP, Computer Vision (CV), and Time Series (TS) domains. However, it is worth noting that most of these methods simply superimpose Transformer modules onto existing cognitive signal decoders, often overlooking the overfitting issue that arises from cognitive signals with limited samples and a low signal-to-noise ratio (SNR), a challenge stemming from the inherent complexities of Transformers.

%
%

In this paper, to address aforementioned issues, we propose D-FaST, a brain cognitive signal decoder that incorporates in a {\bf{D}}sentangled {\bf{F}}requency-{\bf{S}}patial-{\bf{T}}emporal {\bf{A}}ttention(Fig.~\ref{fig_3D}(d)). We extensively explore the application of attention mechanisms in decoding temporal, spatial, and frequency domain information, as well as various frameworks for integrating these three domains. We conduct substantial experiments to validate our approach.

The contributions of this paper can be summarized as follows:

\begin{itemize}
    \item{Designing a disentangled frequency-spatial-temporal structure for EEG processing, which efficiently integrates features from the frequency spatial and temporal domains and avoids mutual interference between orthogonal domains.}
    \item{Introducing an efficient decoding mechanism based on attention mechanisms for frequency, spatial, and temporal domains to capture global dynamic and function-connected feature more effectively, leading to improved EEG information decoding.}
    \item{Conducting extensive experiments on our self-constructed CLP dataset Mandarian Natural Reading EEG dataset (MNRED), as well as Zurich Cognitive Language Processing Corpus (ZuCo) and another two classic CSD datasets. The experimental results demonstrate the effectiveness of our model and achieve state-of-the-art performance. }
\end{itemize}

The remainder of this paper is organized as follows: Section~\ref{related_work} presents the related work. The proposed methodology is illustrated in Section~\ref{methodology}. The performance of D-FaST and the visualization analysis are presented in Section~\ref{experiments}. Finally, Section~\ref{conclution_analysis} summarizes the paper.

\section{Related work}
\label{related_work}

\subsection{Cognitive Language Processing}

Cognitive Signal Decoding (CSD) primarily relied on traditional machine learning techniques such as Support Vector Machines (SVM)~\cite{R77} and Linear Discriminant Analysis (LDA)~\cite{R78}. However, with the demonstrated advantages of CNNs and Recurrent Neural Networks (RNNs), numerous CSD algorithms based on CNNs and RNNs, such as EEGNet~\cite{R1}, ConvNet~\cite{R30}, and ConvLSTM~\cite{R16}, have been designed and continue to play a crucial role in various scenarios. As the field of NLP and CV witnessed the ascension of transformer-based models, several transformer-based CSD algorithms, such as STAGIN~\cite{R18} and TTF-Former~\cite{R69}, have rapidly emerged. Concurrently, multiple datasets have been created to support CSD research~\cite{R47, R74, R76}. For instance, the BraVL multimodal matching dataset~\cite{R46} combines brain, visual, and linguistic data, enabling zero-shot decoding of novel visual categories based on recorded human brain activities through multimodal learning. The ZuCo dataset~\cite{R47} integrates EEG and eye-tracking data, capturing participants' reading of sentences in natural conditions. In this paper, we introduce the first CLP dataset that employs Chinese text as stimulus sources, named MNRED.

\subsection{Frequency feature extraction}

The method for decoding brain cognitive signals primarily employs two approaches for frequency feature extraction. One approach utilizes Time-Frequency Representation (TFR) to express frequency domain information, encompassing techniques such as the smooth pseudo-Wigner-Ville distribution (SPWVD), short-time Fourier transform (STFT), continuous wavelet transform (CWT), and others~\cite{R5, R6, R7, R84}. The second method entails the extraction of frequency information from EEG data through convolution operations. For instance, EEGNet~\cite{R1} employs convolutional kernels to extract features from the frequency domain, with kernel sizes set at half the sampling frequency. Nevertheless, these methods often oversimplify frequency domain features, and their parameter configurations are constrained by human empirical knowledge, thus limiting their efficacy in representing spectral information.
Notably, TimesNet~\cite{R62} transforms 1D time series into a collection of 2D tensors based on multiple periods, fully exploiting the multi-periodicity present in time series data. However, applying such a transformation to cognitive signals poses challenges due to their low signal-to-noise ratio and non-periodic nature. In this paper, we introduce a novel approach for frequency feature extraction, involving the use of multi-view attention.

\subsection{Spatial feature extraction}
\label{spatial_feature_extraction}
Besides the frequency-domain features, spatial characteristics also represent another significant aspect of cognitive signals. Cognitive signals are typically acquired from various brain regions using devices such as EEG caps, inherently containing spatial information through the data represented by distinct channels. These signals exhibit functional connectivity (FC) among different brain regions, often represented as connectivity graphs to encode the spatial correlations between EEG cap nodes or brain regions. BrainNetCNN~\cite{R3} leverages brain connectivity graphs as inputs and models the encoding of cognitive states through convolutions applied to edge-to-edge, edge-to-node, and node-to-graph connections. LMDA-Net~\cite{R4} introduces a channel attention mechanism to assess the significance of different EEG acquisition nodes in encoding cognitive states. Nonetheless, the spatial information within the acquired cognitive signals is inherently two-dimensional, and sometimes even three-dimensional, making simple convolutions less effective for handling complex tasks.
Approaches such as graph-based node arrangement~\cite{R2} mitigate some of the limitations of convolutions by arranging nodes into a two-dimensional layout based on spatial relationships. However, these approaches tend to emphasize anatomical connections while neglecting the functional connectivity (FC) of the brain. Models like BrainGNN~\cite{R8}, IBGNN~\cite{R9}, and TARDGCN~\cite{R88} employ Graph Neural Networks (GNN) to model the FC among brain regions. LOGO~\cite{R81} has also achieved success in multi-variate time series prediction using GNN.
Transformer-based approaches~\cite{R12}, exemplified by BNT~\cite{R11}, encode global features of nodes within brain connectivity graphs and subsequently employ orthogonal clustering methods to compress and extract high-level features.
However, they often primarily consider the static spatial characteristics of brain cognition. In reality, the process of brain cognition is dynamic, with historical states significantly influencing the current cognitive state. Consequently, these approaches struggle to model the dynamic nature of brain cognitive processes, leading to suboptimal utilization of spatial domain information.
In this paper, we introduce a novel dynamic connectogram attention mechanism for the extraction of spatial features in a more dynamic and context-aware manner.

\subsection{Temporal feature extraction}

While the representation of cognitive signals from the orthogonal dimensions of frequency and space domains is sufficiently comprehensive, it is particularly important to investigate the evolution of cognition in the temporal dimension, given that cognitive signals are quintessentially multivariate time series. RNN models~\cite{R13, R14, R15} excel in extracting features from such time series data. ConvLSTM~\cite{R16}, which utilizes Long Short-Term Memory (LSTM)~\cite{R14} to capture dynamic contextual features from brain cognitive signals, is another notable approach. Nonetheless, RNNs face challenges related to parallel computation, leading to heightened computational complexity and making them less suited for the analysis of brain cognitive signals sampled at high rates.

BrainNet~\cite{R17} introduces a self-supervised Bidirectional Contrast Predictive Coding (BCPC) to pretrain a universal feature encoder for brain cognitive signals, effectively addressing the issue of low data utilization stemming from imbalanced EEG data labels. STAGIN~\cite{R18} excels at extracting contextual features from dynamic graphs of brain cognitive signals through the bidirectional encoding capabilities of Transformer structures. However, this approach necessitates a relatively extended EEG data sampling period, and the inclusion of Transformer structures introduces computational complexity, impacting the detailed feature extraction and analytical efficiency of brain cognitive signals.

Numerous research endeavors have focused on enhancing the efficiency of Transformers~\cite{R22, R23, R44, R45}. These studies underscore the effectiveness of attention-based feature extraction in the temporal domain, while acknowledging the imperative need to manage computational costs. EEGNet-MSD~\cite{R71}, which combines EEGNet~\cite{R1} and Informer~\cite{R23}, offers a simple yet potent approach with the potential to enhance cognitive signal decoding performance. EmoGT~\cite{R72} integrates Graph Convolutional Networks (GCN) with Transformer and designs a Cross-modal Attention mechanism to establish connections between EEG data and eye movements. In this paper, we introduce a novel approach: a local temporal sliding attention mechanism designed for the extraction of temporal features.

\subsection{Multidomain feature fusion}
Existing research suggests that spectral, temporal, and spatial information play complementary roles in the analysis of cognitive signals, particularly in interactions between the spatial and spectral domains or the spatial and temporal domains. Consequently, the prevailing approach is to analyze EEG signals using multimodal features from multiple dimensions. This necessitates an efficient feature fusion mechanism for the seamless integration of cross-domain information.
Most popular networks for brain cognitive information analysis adopt a sequential structure in which features from different domains are extracted and analyzed in stages. Notable examples include EEGNet~\cite{R1}, STAGIN~\cite{R18},MSFEnet~\cite{R85}, CDCN~\cite{R86} and FBNetGen~\cite{R24}, among others. BrainNet~\cite{R17} has also developed a spatio-temporal information alternation fusion mechanism based on the diffusion property of EEG. TTF-Former~\cite{R69} incorporates cross-attention to merge temporal and frequency features. However, as previously mentioned, there exists a notable degree of independence between spatial and frequency domain information, as well as between spatial and temporal domains. The sequential processing within staged structures can lead to mutual interference between different domain information during the processing stages, thereby hindering the efficiency of feature extraction.
In this paper, we propose a novel disentangled frequency-spatial-temporal architecture aimed at seamlessly fusing features from all three domains.

\section{Methodology}
\label{methodology}

\begin{figure*}[h!tp]
    \centering
    \includegraphics[width=7in]{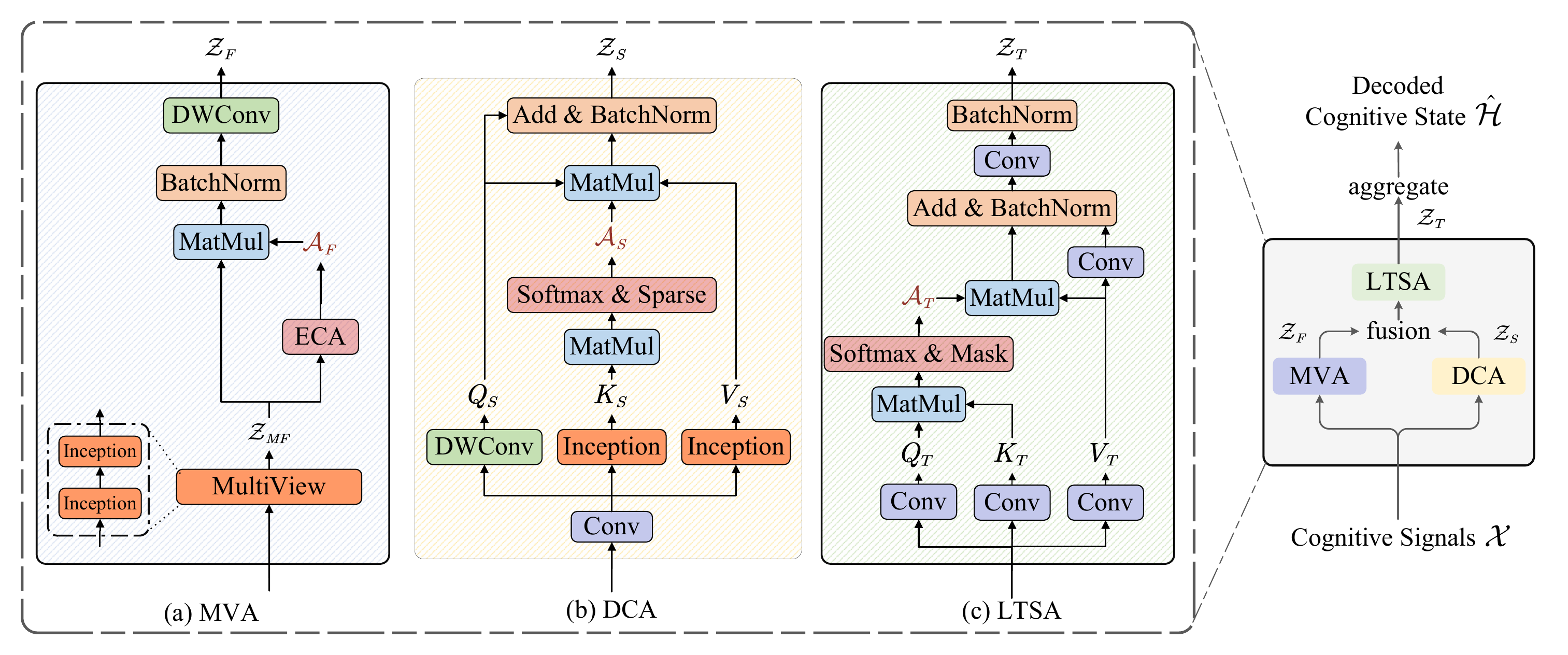}
    \caption{The overarching architecture of D-FaST. The dashed boxes delineate detailed descriptions of the corresponding modules. The three diagrams on the left provide a comprehensive breakdown of the neural networks within the MVA, DCA, and LSTA modules. The rightmost section illustrates the interconnections between these three modules.}
    \label{fig_D-FaST}
\end{figure*}

\subsection{Problem Definition}
\label{problem_definition}

The research goal of CSD is training a brain cognitive decoding network $\mathcal{F}:\mathcal{X}\rightarrow \mathcal{H}$ in which the output $\mathcal{H}\in \mathbb{R}^d$ is a coded representation of cognitive signal $\mathcal{X} \in \mathbb{R}^{N\times T}$.

Given a set of signal acquisition nodes $\mathcal{V} = \{v_1, v_2, \cdots, v_N\}$ distributed in the brain area space, where $N = \lvert \mathcal{V} \rvert$ denotes the number of nodes represented as sensor channels in EEG data, each node samples EEG data at a sampling frequency $f$. The collected brain signal data is expressed as $\mathcal{X} \in \mathbb{R}^{N\times T}$, where $T$ is the number of sampling time points. It is assumed that the labels of brain cognitive tasks are represented as cognitive labels $\mathcal{Y} \in \left\{1, \cdots, C \right\}$, and each brain signal sample in the sample set $\{\mathcal{X}\}$ corresponds to a label. A Multi-Layer Perceptron (MLP) transforms $\hat{\mathcal{H}}$ to logits, where a prediction $\hat{\mathcal{Y}}\in \left\{ 1,\cdots ,C \right\} $ can be acquired.

\begin{algorithm}[h!tp]
    \caption{Pseudo-code of D-FaST.}\label{alg:alg1}
    \label{alg1}
    \begin{algorithmic}
        \STATE \textbf{Require: }Initialized parameters of D-FaST model $\Theta$.
        \STATE \textbf{Require: }Data set of cognitive signals and corresponding labels $\{\mathcal{X}, \mathcal{Y}\}$
        \STATE
        \STATE {\textsc{D-FaST}}$(\mathcal{X})$
        \STATE \hspace{0.5cm}$\mathcal{Z}_{F} \gets MVA(\mathcal{X}) $   \hfill{$\triangleright$Extract frequency feature}
        \STATE \hspace{0.5cm}$\mathcal{Z}_{S} \gets DCA(\mathcal{X})$ \hfill{$\triangleright$Extract spatial feature}
        \STATE \hspace{0.5cm}$\mathcal{Z}_{FS} \gets FUSION(\mathcal{Z}_{F}, \mathcal{Z}_{S}) ^{\top}$ \hfill{$\triangleright$Fuse}
        \STATE \hspace{0.5cm}$\mathcal{Z}_{T} \gets LSTA(\mathcal{Z}_{FS})$ \hfill{$\triangleright$Extract temporal feature}
        \STATE \hspace{0.5cm}$\hat{\mathcal{H}} \gets AGGREGATE( \mathcal{Z}_T )$\ \hfill{$\triangleright$Aggregate}
        \STATE \hspace{0.5cm}\textbf{return}  $\hat{\mathcal{H}}$
        \STATE
        \STATE {\textsc{TRAIN}}$(\{\mathcal{X}, \mathcal{Y}\})$
        \STATE \hspace{0.5cm}$\textbf{for } \mathcal{X}, \mathcal{Y} \in \{\mathcal{X}, \mathcal{Y}\} \textbf{ do }$
        \STATE \hspace{1cm}$\hat{\mathcal{H}} \gets {\textsc{D-FaST}}(\mathcal{X}) $   \hfill{$\triangleright$Forward}
        \STATE \hspace{1cm}$logits \gets  MLP(\hat{\mathcal{H}}) $   \hfill{$\triangleright$Classify}
        \STATE \hspace{1cm}$loss \gets  CROSS\_ENTROPY(logits, \mathcal{Y}) $
        \STATE \hfill{$\triangleright$Calculate loss}
        \STATE \hspace{1cm}$loss.backward()$   \hfill{$\triangleright$Back-propagate}
        \STATE \hspace{1cm}update($\Theta$)  \hfill{$\triangleright$Update parameters using Adam}
        \STATE \hspace{0.5cm}\textbf{end for}
    \end{algorithmic}
\end{algorithm}

One sample data $\mathcal{X}\in \mathbb{R}^{N\times T}$ can be divided into $h$ segments along the temporal axis, with each segment referred to as a time window corresponding to $w$ sampling time points. Taking the $t^{th}$ time window as an example, a connectogram can be constructed as $\mathcal{G}^t=\left\{ \mathcal{V},\mathcal{E}^t \right\}$. $\mathcal{E}^t$ Represents the connection relationship between the brain regions of each sampling node in the $t^{th}$ time window. Such connection relationship is defined using a triplet $\left( v_i,e_{v_iv_j},v_j \right) ,v_i,v_j\in \mathcal{V}$  where the weights of the edges are $e_{v_iv_j}\in \left[ 0,1 \right] $, and $e_{v_iv_j}=e_{v_jv_i}$, indicating that the connection graph described here is an undirected graph. When $e_{v_iv_j}=0$, it indicates no connection between the nodes. Finally, a set of $h$ brain connections $\mathcal{G}=\left\{ \mathcal{G}^t\mid t=1,\cdots ,h \right\} $ is formed.

\subsection{Overview of D-FaST}
In this paper, we introduce a novel network called D-FaST, which aims to enhance the utilization of frequency, spatial, and temporal domains while improving the effectiveness of structure. Fig.~\ref{fig_D-FaST} provides an intuitive overview of D-FaST, while Algorithm~\ref{alg1} describes its overall process using pseudo-code. D-FaST trains a cognitive signal decoder by applying Multi-View Attention (MVA), Dynamic Connectogram Attention (DCA) and Local Temporal Sliding Attention (LTSA). Cognitive signals $\mathcal{X}$ are processed through MVA and DCA, respectively, yielding frequency feature $\mathcal{Z}_{F}$ and spatial feature $\mathcal{Z}_{S}$. LSTA extracts temporal features $\mathcal{Z}_{T}$ from the fused features of $\mathcal{Z}_{F}$ and $\mathcal{Z}_{S}$. $\mathcal{Z}_{T}$ are then aggregated to obtain decoded cognitive state $ \hat{\mathcal{H}}$. D-FaST avoids the mutual interference caused by feature differences between frequency and spatial features by extracting them in a disentangled way.

\subsection{Frequency-Spatial-Temporal Attention}
\label{method_d_FaST}
\subsubsection{\textbf{Multi-View Attention (MVA) for Frequency Feature Extraction}}

\begin{figure}[h!b]
    \centering
    \includegraphics[width=3in]{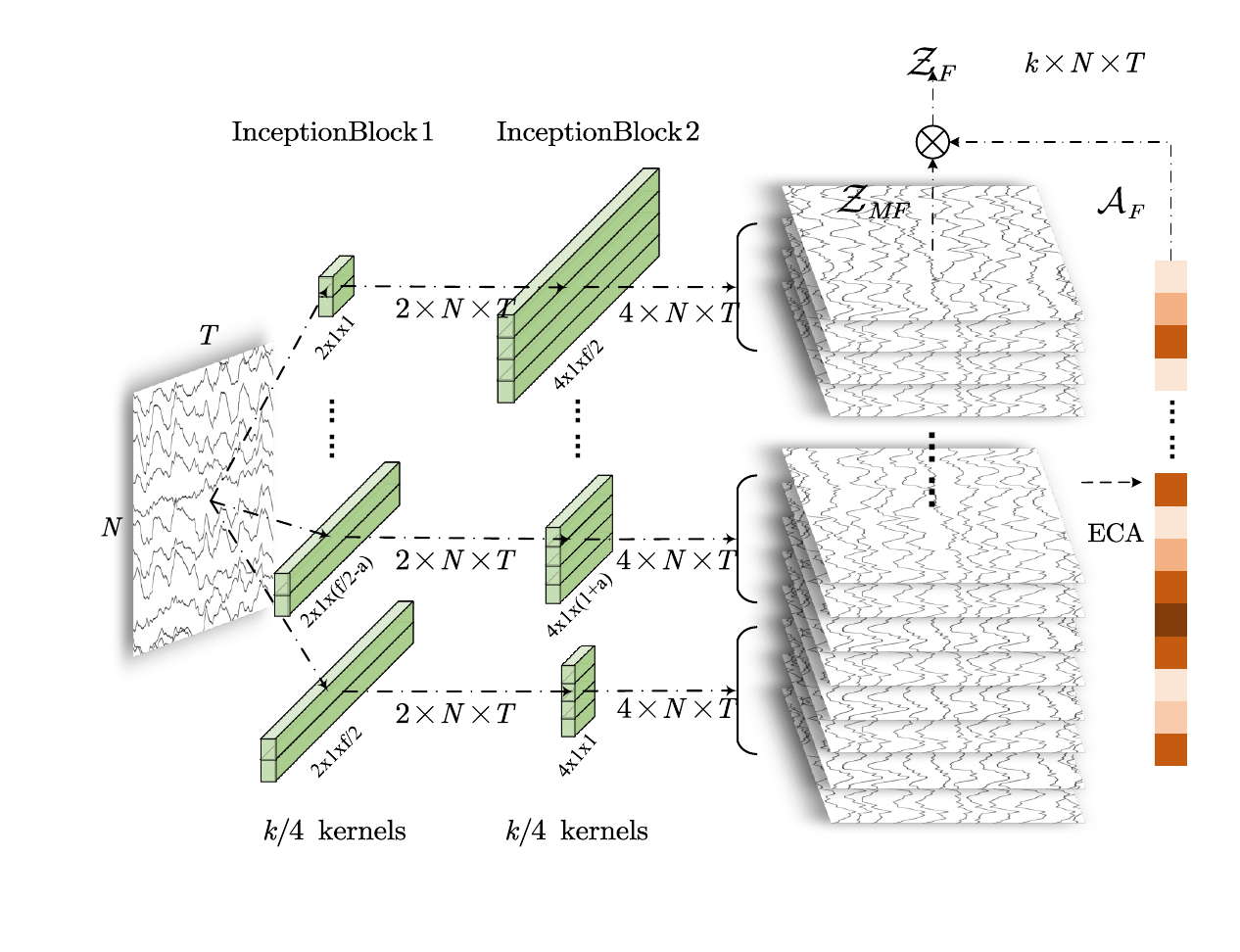}
    \caption{Frequency domain information coding process of multi-view attention.}
    \label{fig_mva}
\end{figure}

Compared to previous methods that relied on single, experiential frequency-domain feature extraction~\cite{R1}, this module focuses on the extraction of non-empirical multi-frequency features and directs the model's attention towards significant frequencies. The feature extraction of brain cognitive signals in the frequency domain is performed using the $\text{MVA} : \mathcal{X} \rightarrow \mathcal{Z}_F \in \mathbb{R}^{k \times N \times T}$, where $k$ denotes the target number of frequency domain features. As illustrated in Fig.~\ref{fig_mva}, MVA consists of two components: a multi-view convolutional structure and frequency attention. The detailed structure of MVA is depicted in Fig.~\ref{fig_D-FaST}(a).

\textbf{Multi-View convolution:}
~\cite{R25} introduced a variety of modular aggregation structures to enhance feature extraction in a disentangled manner. Similarly, multi-view convolution transforms $\mathcal{X}\in \mathbb{R}^{N\times T}$ to $\mathcal{Z}_{MF}\in \mathbb{R}^{k\times N\times T}$. Specifically, multi-view convolution consists of a superposition of two multi-scales InceptionBlocks:
\begin{equation}
    \mathcal{Z}_{MF}=\text{InceptionBlock2}\left( \sigma \text{InceptionBlock1}\left( \mathcal{X} \right) \right)
    \label{mva_1}
\end{equation}
where $\text{InceptionBlock1}:\mathcal{X}\rightarrow \mathcal{Z}_{MF}^{'}\in \mathbb{R}^{\frac{k}{2}\times N\times T}$ consists of $k/4$ groups of evenly spaced convolution kernels ranging from $\left(1,1\right)$ to $\left(\frac{f}{2},1\right)$ with an interval of $a=\left\lfloor\frac{2f}{k}\right\rfloor$. Each group contains two convolution kernels, resulting in the extraction of $\frac{k}{2}$ frequency features in total. $\sigma$ denotes the activation function.

Additionally, $\text{InceptionBlock2}:\mathcal{Z}_{MF}^{'}\rightarrow \mathcal{Z}_{MF}\in \mathbb{R}^{k\times N\times T}$ comprises $k/4$ groups of evenly spaced convolution kernels ranging from $\left(\frac{f}{2},1\right)$ to $\left(1,1\right)$ with an interval of $a=\left\lfloor\frac{2f}{k}\right\rfloor$. Each group convolution includes 4 convolution kernels to extract 4 frequency-domain features from two inputted frequency-domain features. Modifying the convolution kernel sizes enhances the richness and hierarchy of the frequency domain feature extraction process.

\textbf{Multi-View attention:}
To further enhance the quality of multi-frequency features, Multi-View Attention (MVA) assigns attention weights to the extracted frequency features. This approach facilitates a scientific investigation into the significance of various frequency information within brain cognitive signals. Existing models such as SENet~\cite{R26}(Squeeze-and-Excitation), ECA-Net~\cite{R27}(Efficient Channel Attention), and LMDA-Net~\cite{R4} endeavor to elucidate the importance of different channel information through channel attention mechanisms. Similarly, we propose the design of a MVA, denoted as $\text{Attention:}\mathcal{Z}_{MF}\rightarrow \mathcal{Z}_F$, which operates as follows:

\begin{equation}
    \begin{aligned}
        \mathcal{Z}_F= & \text{MVA}\left( \mathcal{X} \right)                                                                     \\
        =              & \text{DWConv}\left( \text{Attention}\left( \mathcal{Z}_{MF} \right) \right)                              \\
        =              & \text{DWConv}\left( \mathcal{A}_F\mathcal{Z}_{MF} \right)                                                \\
        \mathcal{A}_F= & \text{SE}\left( \mathcal{Z}_{MF} \right)                                                                 \\
        =              & \text{Sigmoid}\left( \text{Linear}\left( \text{AvgPool2d}\left( \mathcal{Z}_{MF} \right) \right) \right) \\
    \end{aligned}
    \label{mva_2}
\end{equation}
where $\mathcal{A}_F\in \left[ 0,1 \right] ^k$ represents $k$ attention weights of frequency features of $\mathcal{Z}_{MF}$, and is obtained by applying one-dimensional convolution followed by two-dimensional average pooling with $\mathcal{Z}_{MF}$. The attention weights are then further processed using the sigmoid function. $\text{DWConv}\left( \cdot \right)$ is used to adjust the output dimension of $\mathcal{Z}_F$ in the spatial domain. As mentioned above, the multi-view convolution kernels in the InceptionBlock can be adjusted to capture different frequency ranges. Furthermore, the convolution method is used to obtain local attention, which reduces the computational cost and pays more attention to the relationship between adjacent frequencies. In fact, in order to reduce the training complexity of the model and avoid the overfitting of the model on the data noise, we also add a pooling layer at the end which is omitted from equation (2) for the sake of simplify. Similarly, the subsequent pooling layer is omitted.

\subsubsection{\textbf{Dynamic Connectogram Attention (DCA) for Spatial Feature Extraction}}
\label{DCA}
Compared to previous static spatial feature representation methods that focused on the spatial characteristics of physical nodes~\cite{R3, R4, R11}, this module focuses on the connectivity patterns between virtual regions of interest and their dynamic characterization. The feature extraction of brain cognitive signals in the spatial domain is performed using the $\text{DCA:}\mathcal{X}\rightarrow \mathcal{Z}_S\in \mathbb{R}^{k\times N'\times T}$, where $N'$ is the number of virtual subspace nodes. DCA consists of two parts: dynamic connectogram and multi-head dot-product attention, as shown in Fig.~\ref{fig_D-FaST}(b).

\begin{figure}[h!t]
    \centering
    \includegraphics[width=3in]{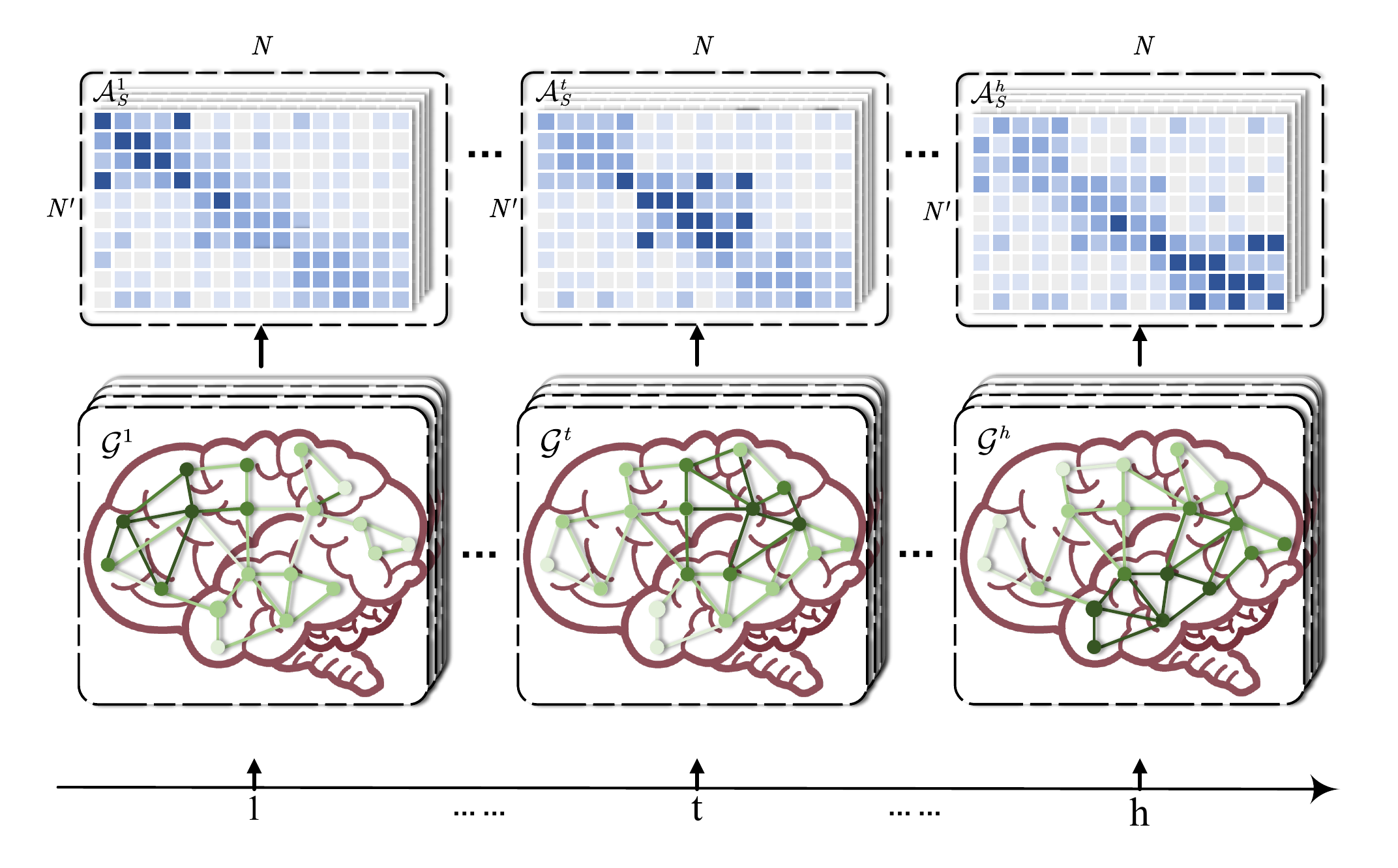}
    \caption{Dynamic connectogram and dynamic connection matrix of each window.}
    \label{fig_dca}
\end{figure}

\textbf{Dynamic Connectogram:} Brain cognitive signals exhibit a natural dynamic graph structure, and the key to various brain functions lies in the connection and communication between different regions~\cite{R28}. In order to fuse with features of other dimensions, DCA first uses one-dimensional convolution to project $\mathcal{X}\in \mathbb{R}^{N\times T}$ to $\mathcal{Z}_{S}^{'}\in \mathbb{R}^{k\times N'\times T}$, $\mathcal{Z}_{S}^{'}=\left\{ \mathcal{Z}_{S}^{'t}\mid t=1,\cdots ,h \right\} $, where $\mathcal{Z}_{S}^{'t}\in \mathbb{R}^{k\times N'\times w}$, $w$ is the size of the sliding window; Then DCA calculated the set of dynamic connection matrices $\mathcal{A}_S=\{\mathcal{A}_{S}^{t}\mid t=1,\cdots ,h\}$ corresponding to the set of dynamic brain connection graphs $\mathcal{G}=\left\{ \mathcal{G}^t\mid t=1,\cdots ,h \right\} $ with $\mathcal{Z}_{S}^{'}$, as shown in Fig.~\ref{fig_dca}, where $\mathcal{A}_{S}^{t}\in \left[ 0,1 \right] ^{k\times N\times N}$ is the connection matrix corresponding to the $t^{th}$ window, calculated as follows:

\begin{equation}
    \mathcal{A}_{S}^{t}=\text{Softmax} \left( \text{Sparse}\left( \frac{{Q_{S}^{t}K_{S}^{t}}^{\top}}{\sqrt{T}},\tau \right) \right)
    \label{dca_1}
\end{equation}
where $\text{DWConv}\left( \cdot \right)$ transforms $\mathcal{Z}_{S}^{'t}$ to subgraph query matrix $Q^t\in \mathbb{R}^{k\times N'\times w}$, and  $\text{InceptionBlock}\left( \cdot \right) $  transforms $\mathcal{Z}_{S}^{'t}$ to key matrix $K^t\in \mathbb{R}^{k\times N\times w}$ . The convolution kernel size used in $\text{DWConv}\left( \cdot \right)$ is $\left( N,1 \right) $; $\tau$ denotes the spatial sparse coefficient. The top $\tau\%$ of the input attention score matrix is retained by $\text{Sparse}\left( \cdot ,\tau \right)$ and the rest $1-\tau \%$ is assigned to be $-\infty$. After activation function $\text{Soft}\max $, the edge with insignificant connection is removed. a scaling operation $1/\sqrt{T}$ is used in the equation to prevent the gradient from disappearing~\cite{R29}.

\textbf{Spatial Context Attention:}
Unlike the multi-head dot-product attention in Transformer~\cite{R29} that operates on the embeddings dimension, DCA performs Spatial Context Attention on the temporal dimension, where dynamic graph features that corresponds to the aforementioned number of Windows are extracted. The specific calculation process is as follows:

\begin{equation}
    \mathcal{Z}_S=\mathcal{A}_SV_S=\sum_{t=1}^h{\mathcal{A}_{S}^{t}V_{S}^{t}}
    \label{dca_2}
\end{equation}

Similar to $Q_{S}^{t}$, $\text{InceptionBlock}\left( \cdot \right) $ transforms $\mathcal{Z}_{S}^{'t}$ to a value matrix $V_{S}^{t}\in \mathbb{R}^{k\times N\times w}$; Summation is used here to aggregate the dynamic information of the sub-graph corresponding to $h$ windows.

\textbf{Virtual Regions of Interest:}
\label{virtual_regions_of_interest}
In the modeling process mentioned above, we noticed that the number of nodes $N'$ in the subgraph query matrix $Q^t\in \mathbb{R}^{k\times N'\times w}$ is defined as the number of nodes in the subspace. When $N'=N$, we can establish the corresponding relationship between the source node and the target node. When $N'\ne N$($N'>N$or$N'<N$), such a correspondence cannot be established. In this case, we can understand $N'$ by the concept of a virtual brain area or virtual node, where $N'$ corresponds to nodes of virtual abstract meaning. The virtual nodes compute the attention-weighted sums of multiple source nodes and can be seen as representing certain categories (in terms of spatial features as connections) of the source nodes. Therefore, they can also be referred to as virtual regions of interest.

\begin{figure}[h!t]
    \centering
    \includegraphics[width=3in]{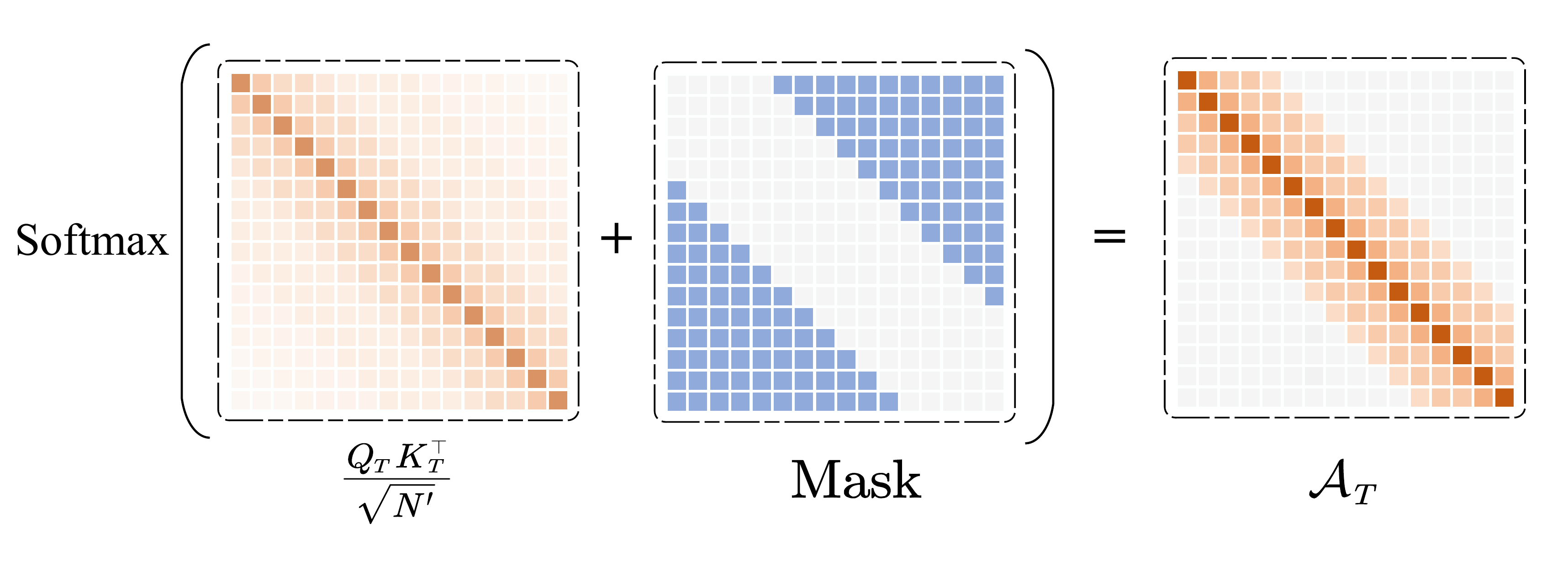}
    \caption{Calculation process of temporal attention weight.}
    \label{fig_lsta}
\end{figure}

\subsubsection{\textbf{Local Temporal Sliding Attention (LTSA)}}

Compared to previous time feature extraction networks with large parameter sizes that disregarded considerations of temporal and spatial complexity~\cite{R16, R18}, which often led to overfitting in small sample scenarios, this module focuses on utilizing lightweight local networks and attention mechanisms to achieve equivalent outcomes. The feature extraction of brain cognitive signals in the temporal domain is performed using the $\text{LTSA:}\mathcal{Z}_{FS}\in \mathbb{R}^{k\times T\times N'}\rightarrow \mathcal{Z}_T\in \mathbb{R}^{k\times T\times N'}$, where $\mathcal{Z}_{FS}=\text{Fusion}\left( \mathcal{Z}_F,\mathcal{Z}_S \right) ^{\top}$ is the fusion of $\mathcal{Z}_F$ and $\mathcal{Z}_S$ in the frequency domain and space, which will be introduced in the next subsection. The LTSA consists of two parts: CNNFormer and local slide-window attention, as depicted in Fig.~\ref{fig_D-FaST}(c).

\textbf{CNNFormer:}
CNNFormer is a Transformer-like model designed for brain cognitive signals. Similar to DCA, LTSA still utilizes dot-product attention in Transformer~\cite{R29}. However, the number of samples of brain cognitive signal data is relatively small. In order to reduce the number of parameters in the network and prevent overfitting, LTSA replaces the method of obtaining query, key and value matrix in Transformer with linear to convolutional operation. Additionally, using convolution allows the preservation of local timing information, whereas using full connections would somewhat disrupt such timing information. The specific equation is calculated as follows:

\begin{equation}
    \begin{aligned}
        \mathcal{Z}_T= & \mathcal{A}_TV_T                                                            \\
        \mathcal{A}_T= & \text{Softmax} \left( \frac{Q_TK_{T}^{\top}}{\sqrt{N'}}+\text{Mask} \right)
    \end{aligned}
    \label{lsta_1}
\end{equation}
where $\mathcal{A}_T\in \mathbb{R}^{k\times T\times T}$ represents the temporal attention score; $\text{CNN}\left( \cdot \right)$ transforms $\mathcal{Z}_{FS}$  respectively to $Q_T,K_T,V_T\in \mathbb{R}^{k\times T\times N'}$; $\text{Mask}\in \mathbb{R}^{T\times T}$ denotes the attention mask, the value of the mask part is $-\infty $, and the remaining unmasked part is a diagonal sliding window of size $w$ with a value of $0$, as shown in Fig.~\ref{fig_lsta}. Furthermore, a scaling operation $1/\sqrt{N'}$ is employed to prevent gradient vanishing.

\textbf{LTSA:}
Despite the relatively small number of samples in brain cognitive signal data, the high sampling frequency in EEG and the long sampling time in fMRI often result in larger samples. If the attention field is not restricted, the network is likely to learn meaningless long-distance contextual semantics while neglecting information at close range. To address this issue, LTSA utilizes local sliding window attention as a means of alleviation. Longformer~\cite{R22} proposed several novel and efficient non-global attention mask mechanisms, achieving favorable outcomes. In this study, we adopt local sliding window attention and present its corresponding mask matrix, as depicted in Figure~\ref{fig_lsta}.

\subsection{Disentangled Frequency-Spatial Feature Extraction}

The existing EEG data processing models can be categorized into three types based on the type and amount of domain information used: single-domain structure, double-domain serial structure, and triple-domain serial structure, as illustrated in Fig.~\ref{fig_3D} (a), (b), and (c), respectively. The single-domain architecture generally only extracts spatial domain information from cognitive signals, with typical models such as BrainNetCNN~\cite{R3} and BNT~\cite{R11}.
The double-domain serial structure primarily extracts both spatial and temporal domain information in different orders. Representative models that extract spatial information first and then temporal information include FBNetGen~\cite{R24} and DeepConvNet~\cite{R30}. Models that extract temporal information first, followed by spatial information, include LAMD-Net~\cite{R4}, STAGIN~\cite{R18}, and ShallowConvNet~\cite{R30}.
The triple-domain serial structure extracts frequency domain information, spatial domain information, and temporal domain information sequentially, with EEGNet~\cite{R1} being the most representative model.

However, these structures fail to capture the differences in relationships between different domains. The use of serial structures for feature extraction across different domains often leads to interference, which affects the effectiveness of feature extraction. To address this, we propose a disentangled frequency-spatial structure, as illustrated in Fig.~\ref{fig_3D}. In this disentangled structure, the frequency domain feature module and the spatial feature module extract features in the frequency and spatial dimensions, respectively, from the cognitive signals. The results are then fused using the temporal module, followed by further aggregation to obtain the cognitive state coding. This can be abstractly expressed as:

\begin{equation}
    \begin{aligned}
        \hat{\mathcal{H}}= & \text{Aggregate}\left( \mathcal{Z}_T \right)                                                                             \\
        \mathcal{Z}_T=     & \text{Temporal}\left( \mathcal{Z}_{FS} \right)                                                                           \\
        \mathcal{Z}_{FS}=  & \text{Fusion}\left( \text{Frequency}\left( \mathcal{X} \right) ,\text{Spatial}\left( \mathcal{X} \right) \right) ^{\top}
    \end{aligned}
\end{equation}
In our approach, the fusion of frequency domain and spatial features is executed in a parallel fashion using $\text{Fusion}\left( \cdot ,\cdot \right)$. This fusion operation can be implemented as either concatenation ($\text{Concat}\left( \cdot ,\cdot \right)$) or addition ($\text{Add}\left( \cdot ,\cdot \right)$). The function $\text{Aggregate}\left( \cdot \right)$ aggregates the tensor $\mathcal{Z}_T$ into a one-dimensional representation. This aggregation can be realized through various techniques such as flattening ($\text{Flatten}\left( \cdot \right)$), mean pooling ($\text{Mean}\left( \cdot \right)$), or employing an attention mechanism ($\text{Attention}\left( \cdot \right)$).
The functions $\text{Temporal}\left( \cdot \right)$, $\text{Temporal}\left( \cdot \right)$, and $\text{Frequency}\left( \cdot \right)$ correspond to $\text{LTSA}\left( \cdot \right)$, $\text{MVA}\left( \cdot \right)$, and $\text{DCA}\left( \cdot \right)$, respectively.


\section{Experimental Evaluation}
\label{experiments}

In this section, we present an evaluation of the effectiveness of our proposed D-FaST model through a comprehensive series of experiments. Our study has been meticulously designed to address the following research questions:

\textbf{Q1.} How does D-FaST perform in comparison to state-of-the-art models featuring various mechanisms and frameworks when applied to CLP dataset?

\textbf{Q2.} How effectively does the model generalize to previous widely-used datasets?

\textbf{Q3.} What is the performance of our proposed components, namely, MVA, DCA, LSTA, and the disentangled framework?

\textbf{Q4.} How do hyperparameters influence the performance?

\textbf{Q5.} To what extent does the trained D-FaST model exhibit interpretability, and how consistent is it with existing knowledge in the field of neuroscience?

\subsection{Datasets and Preprocessing}
We have selected several brain cognitive model datasets that exhibit strong cognitive task correlation. The characteristics of the datasets used in our experiments are summarized in TABLE~\ref{table_dataset}.

\subsubsection*{\bf{MNRED}}

MNRED dataset contains 11,624 EEG signals from 30 native speakers of Mandarin with a gender distribution of 18 males and 12 females, ranged in age from 18 to 25 years. MNRED dataset is a 2-class classification task, and the stimulus materials encompass two categories: target semantic stimuli and non-target semantic stimuli, both in the form of a news headline or a brief sentence. Participants were required to read each stimulus within a 2-second timeframe. EEG data were collected at a sampling rate of 1100 Hz using a 32-channel NeuSen W series wireless EEG acquisition system. Data preprocessing involved referencing to average, resampling the original data to 128 Hz, performing band-pass filtering from 0.1 to 80 Hz, performing independent component analysis (ICA) to remove eye blink and movement artifacts.

\begin{table}[htbp]
    \centering
    \setlength\tabcolsep{2pt}
    \caption{Data set description.}
    \begin{tabular}{c|cccc}
        \toprule
        Dataset         & \multicolumn{1}{c}{MNRED} & \multicolumn{1}{c}{ZuCo}           & \multicolumn{1}{c}{BCIC IV-2a} & \multicolumn{1}{c}{BCIC IV-2b} \\
        \midrule
        Size            & 11624                     & 4478                               & 5184                           & 6520                           \\
        Dimension       & $30\times440$             & $104\times \bigstar\footnotemark $ & $22\times577$                  & $3\times513$                   \\
        Sampling $f$    & 1100Hz                    & 500Hz                              & 250Hz                          & 250Hz                          \\
        Bandpass filter & [0.5,80]                  & [0.5,100]                          & [4,38]                         & [4,38]                         \\
        Subjects        & 10                        & 12                                 & 9                              & 9                              \\
        Classes         & 2                         & 9                                  & 4                              & 2                              \\
        Classes rate    & 3:7                       & 1                                  & 1                              & 1                              \\
        Resampling $f$  & \multicolumn{4}{c}{128Hz}                                                                                                        \\
        \bottomrule
    \end{tabular}
    \label{table_dataset}
\end{table}

\subsubsection*{\bf{ZuCo}}

The ZuCo dataset~\cite{R47} contains eye-tracking and EEG data from 12 participants, all native speakers of English, who performed natural reading and relation extraction tasks on 300 and 407 English sentences from the Wikipedia corpus~\cite{R57}, as well as sentiment reading on 400 samples from the Stanford Sentiment Treebank (SST). We choose the Task-Specific Reading (TSR) task and select EEG signals corresponding to sentences of 10-20 words each. TSR is a ten-class classification task where participants were instructed to attend to a particular type of relation in sentences, including award, education, employer, founder, job title, nationality, political affiliation, visited and wife.

\addtocounter{footnote}{-1}
\addtocounter{footnote}{1}
\footnotetext{$\bigstar$: The sampling lengths in ZuCo are inconsistent and exhibit a large variance, which significantly impacts the data quality when either truncating to a specific length or padding the data.}

\subsubsection*{\bf{BCIC IV-2A}}
The BCI Competition IV Dataset 2A (BCIC IV-2A)~\cite{R31} is a publicly accessible dataset that captures EEG data from 9 subjects participating in motor imagination tasks encompassing four distinct categories: left hand, right hand, foot, and tongue. Data preprocessing procedures involve an initial step of referencing the original data to 128Hz, following the protocol outlined in reference~\cite{R1}. Subsequent steps included band-pass filtering in the frequency range of 4 to 38Hz, followed by a normalization~\cite{R32} and European alignment~\cite{R33}. For model training and testing, two rounds of data were utilized, each comprising approximately 288 records. For each record, the temporal segment following the cue occurrence was extracted, in line with the guidelines presented in references~\cite{R1, R30, R34, R35}.

\subsubsection*{\bf{BCIC IV-2B}}
The BCI Competition IV Dataset 2B (BCIC IV-2B)~\cite{R31} comprises EEG data obtained from 9 subjects participating in two distinct categories of motor imagination tasks involving the left hand and right hand. The data collection procedure and filtering techniques applied are consistent with those employed in BCIC IV-2A. As in the case of BCIC IV-2A, the time segment following the cue occurrence was extracted from each record~\cite{R4}. Subsequently, the data from the 5 rounds for each subject were merged.

\begin{table*}[htbp]
    \centering
    \caption{Compare experimental results under cross-subject experimental settings on MNRED. The optimal results in the table are highlighted in bold, while the suboptimal results are indicated with an underline. This formatting approach is consistently applied in subsequent tables.}
    \begin{tabular}{cccccccc}
        \toprule
        \multirow{2}{*}{Model}                        & \multirow{2}{*}{Venue}                         & \multicolumn{1}{c}{\multirow{2}{*}{Type}}      & \multicolumn{4}{c}{MNRED}                                                                                                \\
        \cline{4-7}
        \multicolumn{1}{c}{}                          & \multicolumn{1}{c}{}                           & \multicolumn{1}{c}{}                           & \makecell[c]{Accuracy (\%)} & \makecell[c]{AUROC(\%)} & \makecell[c]{Sensitivity  (\%)} & \makecell[c]{Specificity (\%)} \\
        \midrule
        \multicolumn{1}{c}{BrainNetCNN~\cite{R3}}     & \scriptsize\textcolor{gray}{[NeuroImage'17]}   & \multicolumn{1}{c}{\multirow{4}{*}{1D}}        & 70.79±0.73                  & 59.46±0.68              & 9.60±0.21                       & \bf{97.17±1.00}                \\
        \multicolumn{1}{c}{BNT~\cite{R11}}            & \scriptsize\textcolor{gray}{[NeurIPS'22]}      & \multicolumn{1}{c}{}                           & 70.76±0.73                  & 61.63±0.69              & 11.45±0.27                      & \underline{96.58±1.02}         \\
        \multicolumn{1}{c}{TACNet~\cite{R34}}         & \scriptsize\textcolor{gray}{[UbiComp'21]}      & \multicolumn{1}{c}{}                           & 74.10±0.78                  & 73.76±0.82              & 51.45±0.70                      & 83.93±0.90                     \\
        \multicolumn{1}{c}{RACNN~\cite{R43}}          & \scriptsize\textcolor{gray}{[IJCAI'21]}        & \multicolumn{1}{c}{}                           & 75.92±0.80                  & 57.74±0.64              & 14.39±0.24                      & 93.81±0.98                     \\
        \midrule
        \multicolumn{1}{c}{DeepConvNet~\cite{R30}}    & \scriptsize\textcolor{gray}{[HBM'17]}          & \multicolumn{1}{c}{\multirow{7}{*}{2D-Serial}} & 74.52±0.79                  & 78.00±0.85              & \bf{69.61±0.79}                 & 76.68±0.83                     \\
        \multicolumn{1}{c}{ShallowConvNet~\cite{R30}} & \scriptsize\textcolor{gray}{[HBM'17]}          & \multicolumn{1}{c}{}                           & 74.19±0.79                  & 71.64±0.82              & 43.30±0.58                      & 87.61±0.92                     \\
        \multicolumn{1}{c}{FBNetGen~\cite{R24}}       & \scriptsize\textcolor{gray}{[MIDL'22]}         & \multicolumn{1}{c}{}                           & 71.83±0.74                  & 65.45±0.72              & 17.58±0.30                      & 95.42±1.00                     \\
        \multicolumn{1}{c}{LMDA-Net~\cite{R4}}        & \scriptsize\textcolor{gray}{[NeuroImage'23]}   & \multicolumn{1}{c}{}                           & 76.00±0.82                  & \underline{78.12±0.88}  & \underline{64.51±0.84}          & 80.98±0.88                     \\
        \multicolumn{1}{c}{EEG-ChannelNet~\cite{R48}} & \scriptsize\textcolor{gray}{[TPAMI'21]}        & \multicolumn{1}{c}{}                           & 74.19±0.78                  & 73.20±0.81              & 48.25±0.63                      & 85.47±0.90                     \\
        \multicolumn{1}{c}{TCACNet~\cite{R42}}        & \scriptsize\textcolor{gray}{[IPM'22]}          & \multicolumn{1}{c}{}                           & 75.92±0.80                  & 74.32±0.84              & 47.32±0.66                      & 88.31±0.93                     \\
        \midrule
        \multicolumn{1}{c}{EEGNet~\cite{R1}}          & \scriptsize\textcolor{gray}{[J Neural Eng'18]} & \multicolumn{1}{c}{3D-Serial}                  & \underline{76.06±0.80}      & 77.94±0.84              & 57.58±0.73                      & 84.13±0.90                     \\
        \midrule
        \multicolumn{1}{c}{\bf{D-FaST}}               & \scriptsize\textcolor{gray}{[\bf{Ours}]}       & \multicolumn{1}{c}{3D-Disentangled}            & \bf{78.72±0.82}             & \bf{81.51±0.85}         & 62.20±0.71                      & 85.98±0.90                     \\
        \bottomrule
    \end{tabular}
    \label{table_MNRED}
\end{table*}

\begin{table}[htbp]
    \centering
    \setlength\tabcolsep{3pt}
    \caption{Compare experimental results under cross-subject experimental settings. Number of baseline models is limited due to the unequal length sampling of the dataset.}
    \begin{tabular}{cccc}
        \toprule
        \multirow{2}{*}{Model}                           & \multirow{2}{*}{Venue}                       & \multicolumn{2}{c}{ZuCo}                              \\
        \cline{3-4}
        \multicolumn{1}{c}{}                             & \multicolumn{1}{c}{}                         & \makecell[c]{Accuracy (\%)} & \makecell[c]{AUROC(\%)} \\
        \midrule
        \multicolumn{1}{c}{FBNetGen~\cite{R24}}          & \scriptsize\textcolor{gray}{[MIDL'22]}       & 76.82±0.80                  & 85.94±1.13              \\
        \multicolumn{1}{c}{BrainNetCNN~\cite{R3}}        & \scriptsize\textcolor{gray}{[NeuroImage'17]} & 76.64±0.78                  & 86.13±1.09              \\
        \multicolumn{1}{c}{Graph Transformer~\cite{R12}} & \scriptsize\textcolor{gray}{[AAAI'21]}       & 77.66±0.81                  & \underline{92.99±0.95}  \\
        \multicolumn{1}{c}{BNT~\cite{R11}}               & \scriptsize\textcolor{gray}{[NeurIPS'22]}    & \underline{77.82±0.81}      & 92.77±0.95              \\
        \midrule
        \multicolumn{1}{c}{\bf{D-FaST}}                  & \scriptsize\textcolor{gray}{[\bf{Ours}]}     & \bf{78.35±0.79}             & \bf{93.19±0.94}         \\
        \bottomrule
    \end{tabular}
    \label{table_ZuCo}
\end{table}

\subsection{Experimental details and evaluation}
The experiment is carried out on a working platform configured with four NVIDIA GeFroce 3090Ti GPUs, and Pytorch is used as the neural network framework. Firstly, the brain cognitive network is randomly initialized and then trained end-to-end in a supervised way based on cross entropy loss.

\subsubsection*{\bf{Baselines}}

Several models are meticulously selected for comparative analysis, including BrainNetCNN~\cite{R3}, BNT~\cite{R11}, DeepConvNet~\cite{R30}, ShallowConvNet, FBNetGen~\cite{R24}, LMDA-Net~\cite{R4}, EEGNet~\cite{R1}, TACNet~\cite{R34}, RACNN~\cite{R43}, EEG-ChannelNet~\cite{R48}, SBLEST~\cite{R41}, and TCACNet~\cite{R42}. It is important to note that the signal collection length of each sample in the ZuCo dataset is not consistent and exhibits a highly random distribution. Many models are incapable of handling variable-length data; therefore, we are only able to evaluate this dataset using FBNetGen~\cite{R24}, BrainNetCNN~\cite{R3}, Graph Transformer~\cite{R12}, and BNT~\cite{R11}.

\subsubsection*{\bf{Evaluation Metrics}}

In the context of a binary task, as exemplified by MNRED, our evaluation function encompasses four key indicators: Accuracy, AUROC (Area Under the Receiver Operating Characteristic curve), Sensitivity, and Specificity. For multi-class classification tasks, typified by BCIC IV-2A, we employ four distinct evaluation measures as our evaluation function, namely Accuracy and the area under the receiver operating characteristic curve (AUROC). The multi-class AUROC, in particular, adopts a one-to-one approach to systematically traverse and average all feasible combinations of classes.

\subsubsection*{\bf{Cross-Subject Setting}}

We conduct leave-one-subject-out cross-validation, and finally reported the mean and standard deviation of experimental performance indicators of all subjects. We also carry out the cross-subject experiment with 5-fold cross-validation using stratified sampling strategy, and the relevant results are in Appendix.~\ref{more_results}.

\subsubsection*{\bf{Within-Subject Setting}}

We perform separate 5-fold cross-validation for each subject, selecting the best value from each fold for evaluation. The mean performance indicators across the five rounds of experiments are calculated for each subject, and the average and variance are reported for the results of the nine subjects.
%
%
\begin{table}[htbp]
    \centering
    \setlength\tabcolsep{2pt}
    \caption{D-FaST hyperparameter setting on different datasets.}
    \begin{tabular}{c|cccc}
        \midrule
        Hyper-Parameter         & MNRED                                  & ZuCo                                  & BCIC IV-2A & BCIC IV-2B \\
        \midrule
        $w$                     & 16                                     & 16                                    & 32         & 3          \\
        $\tau$                  & 0.6                                    & 1                                     & 0.6        & 1          \\
        $N'$                    & 30                                     & 104                                   & 22         & 1          \\
        mini batch size         & 16                                     & 1                                     & 32         & 32         \\
        epochs                  & 200                                    & 100                                   & 200        & 200        \\
        Dropout                 & 0.1                                    & 0.5                                   & 0.5        & 0.5        \\
        learning rate           & \multicolumn{2}{c}{0.0001$\to$0.00001} & \multicolumn{2}{c}{0.001$\to$0.00001}                           \\
        $k$                     & \multicolumn{4}{c}{64}                                                                                   \\
        $h$                     & \multicolumn{4}{c}{4}                                                                                    \\
        weight decay~\cite{R38} & \multicolumn{4}{c}{0.0001}                                                                               \\
        activation              & \multicolumn{4}{c}{GeLU}                                                                                 \\
        normalization           & \multicolumn{4}{c}{BatchNorm}                                                                            \\
        schedule                & \multicolumn{4}{c}{Cosine~\cite{R37}}                                                                    \\
        optimizer               & \multicolumn{4}{c}{Adam~\cite{R36}}                                                                      \\
        \bottomrule
    \end{tabular}
    \label{table_hyperparameter_DFaST}
\end{table}
\begin{table*}[htbp]
    \centering
    \setlength\tabcolsep{2.5pt}
    \caption{Compare experimental results under cross-subject experimental settings.}
    \begin{tabular}{ccccccccccc}
        \toprule
        \multirow{2}{*}{Model}                        & \multirow{2}{*}{Venue}                         & \multicolumn{1}{c}{\multirow{2}{*}{Type}}      & \multicolumn{2}{c}{BCIC IV-2A} &                           & \multicolumn{4}{c}{BCIC IV-2B}                                                                                                                              \\
        \cline{4-5} \cline{7-10}
        \multicolumn{1}{c}{}                          & \multicolumn{1}{c}{}                           & \multicolumn{1}{c}{}                           & \makecell[c]{Accuracy (\%)}    & \makecell[c]{AUROC  (\%)} &                                & \makecell[cc]{Accuracy (\%)} & \makecell[cc]{AUROC(\%)} & \makecell[c]{Sensitivity  (\%)} & \makecell[c]{Specificity (\%)} \\
        \midrule
        \multicolumn{1}{c}{BrainNetCNN~\cite{R3}}     & \scriptsize\textcolor{gray}{[NeuroImage'17]}   & \multicolumn{1}{c}{\multirow{4}{*}{1D}}        & 35.11±0.41                     & 63.01±0.70                             &                                & -                            & -                        & -                               & -                              \\
        \multicolumn{1}{c}{BNT~\cite{R11}}            & \scriptsize\textcolor{gray}{[NeurIPS'22]}      & \multicolumn{1}{c}{}                           & 33.83±0.41                     & 60.59±0.70                &                                & -                            & -                        & -                               & -                              \\
        \multicolumn{1}{c}{TACNet~\cite{R34}}         & \scriptsize\textcolor{gray}{[UbiComp'21]}      & \multicolumn{1}{c}{}                           & 50.33±0.66                     & 72.35±0.86                &                                & 74.05±0.84                   & 81.00±0.93               & 74.23±0.92                      & 73.87±0.91                     \\
        \multicolumn{1}{c}{RACNN~\cite{R43}}          & \scriptsize\textcolor{gray}{[IJCAI'21]}        & \multicolumn{1}{c}{}                           & 38.39±0.45                     & 63.18±0.71                &                                & 72.54±0.83                   & 78.44±0.91               & 69.04±0.80                      & 76.04±0.88                     \\
        \midrule
        \multicolumn{1}{c}{DeepConvNet~\cite{R30}}    & \scriptsize\textcolor{gray}{[HBM'17]}          & \multicolumn{1}{c}{\multirow{7}{*}{2D-Serial}} & \underline{54.73±0.66}         & \bf{79.17±0.87}           &                                & 74.61±0.83                   & 81.83±0.92               & \underline{76.10±0.86}          & 73.12±0.86                     \\
        \multicolumn{1}{c}{ShallowConvNet~\cite{R30}} & \scriptsize\textcolor{gray}{[HBM'17]}          & \multicolumn{1}{c}{}                           & 50.42±0.65                     & 73.72±0.87                &                                & 69.42±0.84                   & 73.41±0.93               & 61.84±0.79                      & \underline{77.00±0.93}         \\
        \multicolumn{1}{c}{FBNetGen~\cite{R24}}       & \scriptsize\textcolor{gray}{[MIDL'22]}         & \multicolumn{1}{c}{}                           & 36.36±0.45                     & 64.27±0.73                &                                & 53.41±0.60                   & 53.89±0.62               & 64.01±1.06                      & 42.81±0.86                     \\
        \multicolumn{1}{c}{LMDA-Net~\cite{R4}}        & \scriptsize\textcolor{gray}{[NeuroImage'23]}   & \multicolumn{1}{c}{}                           & 52.24±0.67                     & 74.10±0.87                &                                & \underline{74.81±0.83}       & \underline{82.27±0.93}   & \bf{77.80±0.85}                 & 71.82±0.87                     \\
        \multicolumn{1}{c}{EEG-ChannelNet~\cite{R48}} & \scriptsize\textcolor{gray}{[TPAMI'21]}        & \multicolumn{1}{c}{}                           & 47.74±0.61                     & 71.95±0.84                &                                & 70.85±0.79                   & 78.71±0.90               & 68.56±0.86                      & 73.13±0.89                     \\
        \multicolumn{1}{c}{SBLEST~\cite{R41}}         & \scriptsize\textcolor{gray}{[TPAMI'23]}        & \multicolumn{1}{c}{}                           & -                              & -                         &                                & 67.68±0.09                   & 76.58±0.13               & 68.70±0.13                      & 66.57±0.22                     \\
        \multicolumn{1}{c}{TCACNet~\cite{R42}}        & \scriptsize\textcolor{gray}{[IPM'22]}          & \multicolumn{1}{c}{}                           & 51.33±0.66                     & 73.22±0.87                &                                & 73.81±0.83                   & 80.12±0.92               & 74.57±0.87                      & 73.05±0.87                     \\
        \midrule
        \multicolumn{1}{c}{EEGNet~\cite{R1}}          & \scriptsize\textcolor{gray}{[J Neural Eng'18]} & \multicolumn{1}{c}{3D-Serial}                  & 53.59±0.72                     & 74.62±0.90                &                                & 73.17±0.82                   & 81.79±0.95               & 70.21±0.90                      & 76.14±0.91                     \\
        \midrule
        \multicolumn{1}{c}{\bf{D-FaST}}               & \scriptsize\textcolor{gray}{[\bf{Ours}]}       & \multicolumn{1}{c}{3D-Disentangled}            & \bf{54.96+0.71}                & \underline{74.48±0.88}    &                                & \bf{76.81±0.77}              & \bf{83.99±0.84}          & 73.89±0.74                      & \bf{79.72±0.80}                \\
        \bottomrule
    \end{tabular}
    \label{table_cross}
\end{table*}
\begin{table*}[htbp]
    \centering
    \setlength\tabcolsep{2.5pt}
    \caption{Compare experimental results under within-subject experimental setting.}
    \begin{tabular}{cccccccccc}
        \toprule
        \multirow{2}{*}{Model}                        & \multirow{2}{*}{Venue}                         & \multicolumn{1}{c}{\multirow{2}{*}{Type}}      & \multicolumn{2}{c}{BCIC IV-2A} &                         & \multicolumn{4}{c}{BCIC IV-2B}                                                                                                                            \\
        \cline{4-5} \cline{7-10}
        \multicolumn{1}{c}{}                          & \multicolumn{1}{c}{}                           & \multicolumn{1}{c}{}                           & \makecell[c]{Accuracy (\%)}    & \makecell[c]{AUROC(\%)} &                                & \makecell[c]{Accuracy (\%)} & \makecell[c]{AUROC(\%)} & \makecell[c]{Sensitivity  (\%)} & \makecell[c]{Specificity (\%)} \\
        \midrule
        \multicolumn{1}{c}{BrainNetCNN~\cite{R3}}     & \scriptsize\textcolor{gray}{[NeuroImage'17]}   & \multicolumn{1}{c}{\multirow{4}{*}{1D}}        & 62.52±11.64                    & 80.66±9.35              &                                & 61.55±5.95                  & 62.15±7.56              & 61.09±8.57                      & 62.01±12.5                     \\
        \multicolumn{1}{c}{BNT~\cite{R11}}            & \scriptsize\textcolor{gray}{[NeurIPS'22]}      & \multicolumn{1}{c}{}                           & 64.91±13.19                    & 82.07±10.38             &                                & 60.27±7.10                  & 62.11±9.22              & 56.39±15.2                      & 64.15±12.3                     \\
        \multicolumn{1}{c}{TACNet~\cite{R34}}         & \scriptsize\textcolor{gray}{[UbiComp'21]}      & \multicolumn{1}{c}{}                           & 74.62±16.28                    & 88.18±10.90             &                                & 81.69±0.96                  & 85.89±1.03              & 79.68±0.95                      & 83.71±1.00                     \\
        \multicolumn{1}{c}{RACNN~\cite{R43}}          & \scriptsize\textcolor{gray}{[IJCAI'21]}        & \multicolumn{1}{c}{}                           & -                              & -                       &                                & 68.45±14.0                  & 70.20±16.8              & 71.40±13.3                      & 65.50±19.7                     \\
        \midrule
        \multicolumn{1}{c}{DeepConvNet~\cite{R30}}    & \scriptsize\textcolor{gray}{[HBM'17]}          & \multicolumn{1}{c}{\multirow{7}{*}{2D-Serial}} & 72.24±14.47                    & 88.10±10.46             &                                & 80.45±11.9                  & 85.92±12.3              & \bf{82.26±9.26}                 & 78.64±15.8                     \\
        \multicolumn{1}{c}{ShallowConvNet~\cite{R30}} & \scriptsize\textcolor{gray}{[HBM'17]}          & \multicolumn{1}{c}{}                           & \underline{81.69±12.89}        & \bf{93.24±7.12}         &                                & 79.03±14.3                  & 83.64±16.6              & 75.20±18.2                      & 82.87±13.7                     \\
        \multicolumn{1}{c}{FBNetGen~\cite{R24}}       & \scriptsize\textcolor{gray}{[MIDL'22]}         & \multicolumn{1}{c}{}                           & 72.78±15.93                    & 87.44±10.52             &                                & 71.27±13.7                  & 74.04±17.3              & 71.73±14.3                      & 70.82±14.7                     \\
        \multicolumn{1}{c}{LMDA-Net~\cite{R4}}        & \scriptsize\textcolor{gray}{[NeuroImage'23]}   & \multicolumn{1}{c}{}                           & 75.29±17.46                    & 88.91±11.44             &                                & 81.23±13.4                  & 86.33±15.1              & 77.72±17.4                      & \underline{84.74±14.1}         \\
        \multicolumn{1}{c}{EEG-ChannelNet~\cite{R48}} & \scriptsize\textcolor{gray}{[TPAMI'21]}        & \multicolumn{1}{c}{}                           & -                              & -                       &                                & 74.39±0.89                  & 80.75±0.97              & 70.15±0.92                      & 78.63±0.99                     \\
        \multicolumn{1}{c}{SBLEST~\cite{R41}}         & \scriptsize\textcolor{gray}{[TPAMI'23]}        & \multicolumn{1}{c}{}                           & -                              & -                       &                                & 76.45±13.9                  & 84.43±15.8              & 74.57±14.2                      & 78.57±20.7                     \\
        \multicolumn{1}{c}{TCACNet~\cite{R42}}        & \scriptsize\textcolor{gray}{[IPM'22]}          & \multicolumn{1}{c}{}                           & 75.20±15.59                    & 88.84±10.30             &                                & 81.21±14.9                  & 85.60±17.2              & 78.34±16.7                      & 84.07±14.8                     \\
        \midrule
        \multicolumn{1}{c}{EEGNet~\cite{R1}}          & \scriptsize\textcolor{gray}{[J Neural Eng'18]} & \multicolumn{1}{c}{3D-Serial}                  & 81.23±15.65                    & 92.37±8.56              &                                & \underline{82.60±14.4}      & \bf{87.55±15.3}         & \underline{81.67±16.2}          & 83.54±16.8                     \\
        \midrule
        \multicolumn{1}{c}{\bf{D-FaST}}               & \scriptsize\textcolor{gray}{[\bf{Ours}]}       & \multicolumn{1}{c}{3D-Disentangled}            & \bf{83.08±13.86}               & \underline{92.92±7.42}  &                                & \bf{83.15±14.2}             & \underline{87.29±15.9}  & 79.80±19.8                      & \bf{86.51±10.5}                \\
        \bottomrule
    \end{tabular}
    \label{table_within}
\end{table*}
\subsection{Performance on CLP datasets (Q1) }
\label{experiment_MNRED}

We conducted cross-subject and within-subject cognitive classification experiments on MNRED and ZuCo respectively.

\subsubsection*{\bf{Experimental Settings}}

Hyperparameter settings of D-FaST are summarized in TABLE~\ref{table_hyperparameter_DFaST} and that of compared models are summarized in Appendix.~\ref{hyperparameter_baseline}. For ZuCo dataset, we remove the LSTA module and only utilized the DCA module for feature extraction due to the variable length of samples in the dataset. The resulting spatial features are then flattened and input into a MLP. To fairly compare model performance, all models use the same optimizer, learning rate and schedule, minibatch size and number of iterations, and weight decay absorption.

\subsubsection*{\bf{Results}}

As shown in TABLE~\ref{table_MNRED}, TABLE~\ref{table_ZuCo}, the method D-FaST that we designed achieved an average accuracy of 78.72\% and an AUROC of 81.51\% in leave-one-subject-out cross-validation on binary classification dataset MNRED, an average accuracy of 78.35\% in within-subject on experiment on 9-class classification dataset ZuCo. The results show that the effect of D-FaST is far superior to other models. Interestingly, models that use only spatial domain features, BrainNetCNN~\cite{R3} and BNT~\cite{R11}, perform poorly on the MNRED dataset. We believe that EEG data have lower spatial resolution than fMRI data, and that such models ignore cognitive processes in time and useful information in the frequency domain and use only limited spatial features. It is worth mentioning that in order to compare these models more fairly, we conducted experiments under different frequency domain feature number settings, as depicted in Fig.~\ref{fig3}(a). The results show that D-FaST has advantages under different frequency domain feature number settings. Fig.~\ref{fig_t_sne_mnred} visualizes D-FaST's significant discriminant properties in decoding MNRED.

\subsection{Generalization ability on traditional datasets (Q2)}
We verify the generalization ability of D-FaST against baseline models on traditional CSD datasets BCIC IV-2A and BCIC IV-2B under different setting: Cross-Subject and Within-Subject.

\subsubsection*{\bf{Experimental Settings}}

Detailed hyperparameter settings for D-FaST and baseline models can be found in TABLE~\ref{table_hyperparameter_DFaST} and Appendix.~\ref{hyperparameter_baseline}.

\subsubsection*{\bf{Results}}
As shown in TABLE~\ref{table_cross}, in the cross-subject experiment on the BCIC IV-2A and BCIC IV-2B datasets. D-FaST achieves an average accuracy of 54.96\%(+0.23\%) and 74.48\% AUROC on dataset BCIC IV-2A. D-FaST also pushes average accuracy and AUROC on BCIC IV-2B to 76.81\%(+2.20\%) and 83.99\%(+1.73\%) with Sensitivity being 73.89\% and Specificity being 79.72\%(+2.72\%).

As shown in TABLE~\ref{table_within}, in the within-subject experiment on the BCIC IV-2A,  D-FaST achieves an average accuracy of 83.08\%(+1.85\%) and an AUROC of 92.92\%. On the BCIC IV-2B dataset, D-FaST achieved an average accuracy of 83.15\%(+0.55\%). These results outperformed other models significantly. More experimental results about each subject can be found in TABLE~\ref{table_within_2A} and TABLE~\ref{table_within_2B}.

Additionally, D-FaST showed the second lowest standard deviations in accuracy (13.86) and AUROC (7.42) when evaluated on the nine subjects, indicating its higher stability. In contrast, LMDA-Net~\cite{R4} and TACNet~\cite{R34}, while potentially achieving optimal performance on specific test set partitions, lack stability and perform poorly in cross-validation. It is worth mentioning that increasing the parameter size of EEGNet improved accuracy on the MNRED dataset but had the opposite effect on BCIC IV-2A and BCIC IV-2B datasets. We believe that the BCIC IV-2A dataset is comparatively easier than MNRED, and the overall low accuracy may be due to a low signal-to-noise ratio and poor data quality in some subjects. Consequently, increasing the parameter size of EEGNet would cause the model to learn noise and overfit. In contrast, D-FaST exhibits inherent robustness, as demonstrated in subsequent hyperparameter sensitivity experiments in Section~\ref{hyperparameter_sensitivity}, which helps mitigate overfitting to some extent.


\subsection{Ablation Study (Q3)}

Ablation experiments are carried out for frequency-time-space improvement and the design of disentangled framework, with EEGNet-large as the baseline. The experiment carried out a 5-fold cross-validation on MNRED and report the average accuracy and the standard deviation.

\subsubsection{Performance Improved By Frequency-Temporal-Spatial Attention}
Experiments are carried out on D-FaST using only one module improvement in frequency, temporal or spatial dimension. The results in TABLE~\ref{table_ablation} indicate that the model performs better than the baseline when any of the three improvements are used alone. The performance of combining the three modules with the disentangled framework is not only better than the experimental setup of using the three modules alone, but better than combining them with serial structure.

\begin{table}[h!tp]
    \centering
    \caption{The impact of MFA, DSA, CTA, and disentangled frameworks on the model.}
    \begin{tabular}{ccccc|c}
        \toprule
        \multicolumn{1}{c}{Method}   & \multicolumn{1}{c}{MFA} & \multicolumn{1}{c}{DCA} & \multicolumn{1}{c}{LSTA} & \multicolumn{1}{c}{Disentangled} & \multicolumn{1}{c}{Accuracy (\%) } \\
        \midrule
        \multicolumn{1}{c}{Baseline} &                         &                         &                          &                                  & 76.06±0.80                         \\
        \midrule
        \multirow{5}{*}{D-FaST}      & \checkmark              & \checkmark              & \checkmark               & \checkmark                       & \bf{\underline{78.72±0.82}}        \\
        \multicolumn{1}{c}{}         & \checkmark              & \checkmark              & \checkmark               &                                  & \bf{77.64±0.82}                    \\
        \multicolumn{1}{c}{}         & \checkmark              &                         &                          &                                  & \bf{76.77±0.81}                    \\
        \multicolumn{1}{c}{}         &                         & \checkmark              &                          &                                  & \bf{77.10±0.82}                    \\
        \multicolumn{1}{c}{}         &                         &                         & \checkmark               &                                  & \bf{76.67±0.81}                    \\
        \bottomrule
    \end{tabular}
    \label{table_ablation}
\end{table}

\subsubsection{Rationality of Disentangled Framework}
The disentangled framework and serial framework are compared on D-FaST and EEGNet-large, respectively. The experiments show that the disentangled framework performs better on D-FaST and produces similar results to EEGNet-large. This indicates that frequency and space are not necessarily progressive relations, and the serial framework may not be the best combination of frequency and space. The disentangled framework can more fully integrate the two characteristics.

\subsubsection{Effect of Fusion Method on Performance}
The fusion methods, $\text{Concat}\left( \cdot ,\cdot \right)$ makes the frequency domain dimension features and spatial dimension features side by side, while $\text{Add}\left( \cdot ,\cdot \right) $ superimpose them. The main difference between the two methods is their gradient backpasses during training. The performance of these two functions on the MNRED dataset is compared. As shown in TABLE~\ref{table_fusion}, the use of superposition is advantageous for this dataset. Although more complex fusion mechanisms could be designed, previous studies have found that splicing and direct overlay are usually the most cost-effective ways without adding a large number of additional parameters~\cite{R18, R40}.

\begin{table}[htbp]
    \centering
    \caption{Effect of different fusion methods on model performance on MNRED dataset.}
    \begin{tabular}{c|cccc}
        \toprule
        \multicolumn{1}{c}{\makecell[c]{Fusion                                                             \\ Method}} & \makecell[c]{Accuracy} & \makecell[c]{AUROC} & \makecell[c]{Sensitivity} & \makecell[c]{Specificity} \\
        \midrule
        \multicolumn{1}{c}{Concat} & 77.01±0.81      & 78.47±0.86      & 56.38±0.73      & 85.99±0.90      \\
        \multicolumn{1}{c}{Add}    & \bf{78.72±0.82} & \bf{81.51±0.85} & \bf{62.20±0.71} & \bf{85.98±0.90} \\
        \bottomrule
    \end{tabular}
    \label{table_fusion}
\end{table}
\begin{table}[htbp]
    \centering
    \caption{Effect of different aggregate methods on model performance on MNRED dataset.}
    \begin{tabular}{c|cccc}
        \toprule
        \multicolumn{1}{c}{\makecell[c]{Aggregate                                                             \\ Method}} & \makecell[c]{Accuracy} & \makecell[c]{AUROC} & \makecell[c]{Sensitivity} & \makecell[c]{Specificity} \\
        \midrule
        \multicolumn{1}{c}{Flatten}   & \bf{78.72±0.82} & \bf{81.51±0.85} & \bf{62.20±0.71} & 85.98±0.90      \\
        \multicolumn{1}{c}{Mean}      & 76.59±0.81      & 77.68±0.85      & 54.24±0.70      & \bf{86.32±0.91} \\
        \multicolumn{1}{c}{Attention} & 76.56±0.81      & 77.82±0.85      & 56.34±0.74      & 85.35±0.90      \\
        \bottomrule
    \end{tabular}
    \label{table_aggregate}
\end{table}

\subsubsection{Effect of Aggregate Method on Performance}
In order to facilitate the final state to the projection function, the aggregation function transforms the two-dimensional feature from the time dimension into an one-dimensional vector. There are three different approaches used: $\text{Flatten}\left( \cdot \right) $, $\text{Mean}\left( \cdot \right)$ and $\text{Attention}\left( \cdot \right)$~\cite{R18}. We tested the performance of these three aggregate functions on the MNRED dataset and found that the flatten approach exhibited superior performance, as depicted in TABLE~\ref{table_aggregate}.

\subsection{Hyperparameter Sensitivity (Q4)}
\label{hyperparameter_sensitivity}

\begin{figure*}[htbp]
    \centering
    \subfloat[]{\includegraphics[width=0.23\linewidth]{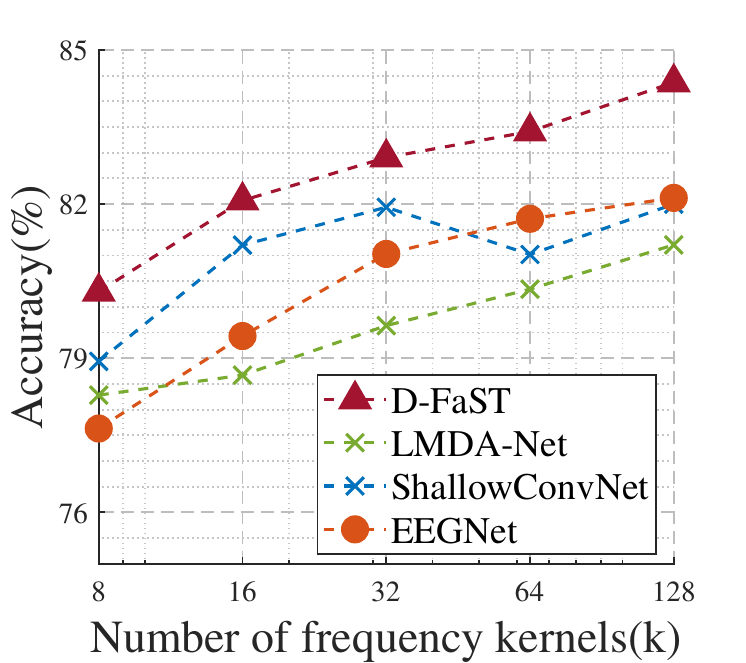}}
    \subfloat[]{\includegraphics[width=0.23\linewidth]{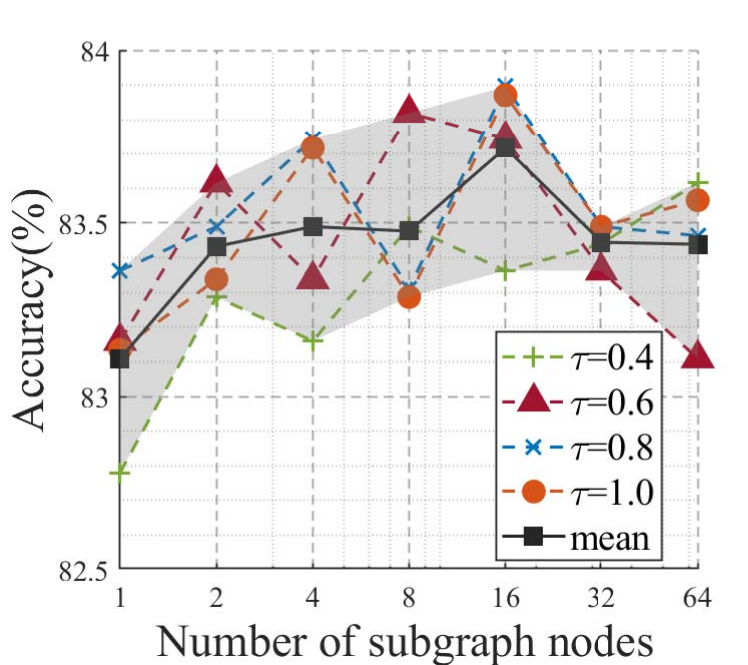}}
    \subfloat[]{\includegraphics[width=0.23\linewidth]{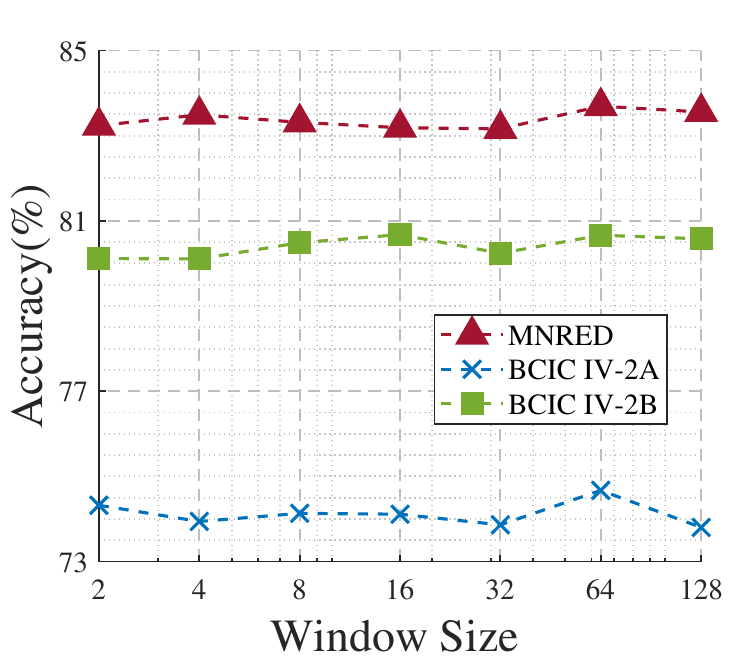}}
    \subfloat[]{\includegraphics[width=0.28\linewidth]{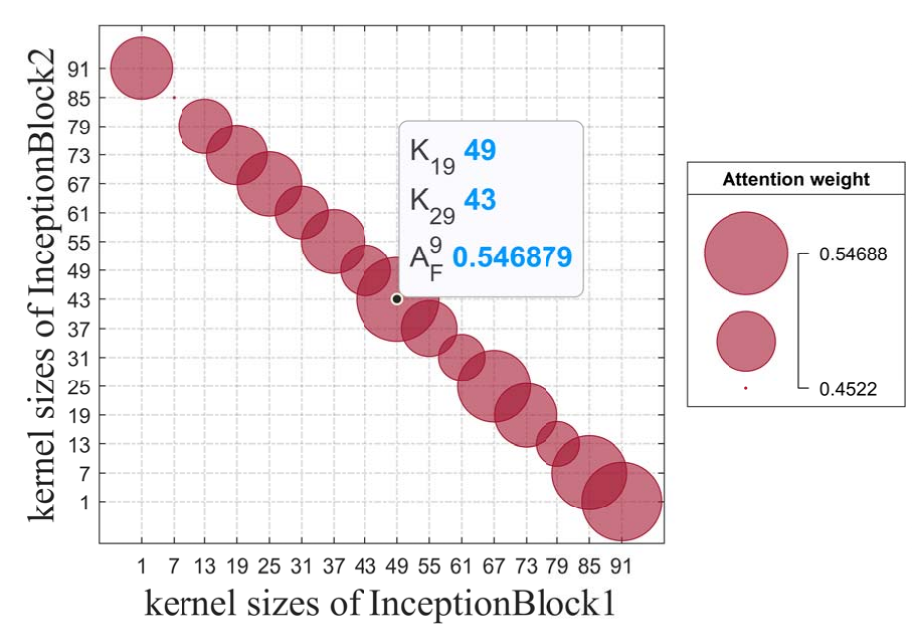}}
    \caption{Dynamic connectogram of negative and positive samples.  (a): Performance difference of D-FaST in different frequency domain feature number Settings. (b):  Performance difference of D-FaST under the setting of spatial sparsity coefficient and number of subspace nodes. The dashed lines of different colors represent the accuracy changes of different spatial sparsity under each subspace node. The solid black line is the mean value of all the dashed lines. The gray area is the range of accuracy for all spatial sparsity and all subspace node configurations. (c):Performance differences of D-FaST under different window size Settings. (d):Attention Weight of Multi-view. The highlighted bubble indicates the largest attention weight, there $K_{19}$ denote the $9^{th}$ convolution kernel size of $\text{InceptionBlock1}$, $K_{29}$ denote the $9^{th}$ convolution kernel size of $\text{InceptionBlock2}$ and $\mathcal{A}_F^9$ represents the attention weight of the $9^{th}$ view.}
    \label{fig3}
\end{figure*}

\begin{table*}[htbp]
    \centering
    \setlength\tabcolsep{2pt}
    \caption{ Performance of within subject experiment with different models on BCIC IV-2A.}
    \begin{tabular}{ccccccccccc}
        \toprule
        \multirow{2}{*}{Model}    & \multirow{2}{*}{Venue}                         & \multicolumn{9}{c}{BCIC IV-2A}                                                                                                                                                                                                         \\
        \cline{3-11}
        \multicolumn{1}{c}{}      & \multicolumn{1}{c}{}                           & A01                            & A02                    & A03                    & A04                    & A05                    & A06                    & A07                    & A08                    & A09                    \\
        \midrule
        BrainNetCNN~\cite{R3}     & \scriptsize{\textcolor{gray}{[NeuroImage'17]}} & 69.44±5.48                     & 53.82±3.70             & 77.43±4.13             & 54.18±7.78             & 43.92±2.54             & 52.95±3.40             & 65.27±6.41             & 75.18±2.90             & 70.49±5.69             \\
        BNT~\cite{R11}            & \scriptsize\textcolor{gray}{[NeurIPS'22]}      & 75.16±4.34                     & 52.60±4.62             & 77.43±2.13             & 60.08±5.64             & 44.61±4.37             & 50.00±4.54             & 71.53±0.79             & 79.00±1.74             & 73.79±2.81             \\
        TACNet~\cite{R34}         & \scriptsize\textcolor{gray}{[UbiComp'21]}      & 81.08±3.98                     & 58.51±2.02             & 90.10±3.05             & 71.87±3.21             & 50.88±5.47             & 53.81±5.19             & 90.10±3.35             & 87.67±3.33             & 87.51±2.67             \\
        DeepConvNet~\cite{R30}    & \scriptsize\textcolor{gray}{[HBM'17]}          & 78.30±2.20                     & 66.31±4.37             & 82.45±5.33             & 76.38±3.90             & 51.73±2.21             & 45.84±5.33             & 80.19±7.84             & 83.50±4.78             & 85.42±2.33             \\
        ShallowConvNet~\cite{R30} & \scriptsize\textcolor{gray}{[HBM'17]}          & 88.38±2.76                     & \bf{68.22±4.28}        & 90.97±2.51             & \underline{83.00±4.65} & \bf{69.09±2.45}        & 59.19±5.81             & \bf{95.84±1.42}        & 91.49±1.12             & 89.07±3.15             \\
        LMDA-Net~\cite{R4}        & \scriptsize\textcolor{gray}{[NeuroImage'23]}   & 82.29±2.01                     & 64.06±5.77             & 93.57±1.91             & 69.09±5.12             & 48.61±3.72             & 51.38±5.90             & 89.06±3.62             & 89.23±3.30             & 90.28±1.15             \\
        FBNetGen~\cite{R24}       & \scriptsize\textcolor{gray}{[MIDL'22]}         & 78.47±5.07                     & 61.81±2.27             & 86.80±1.45             & 60.60±4.09             & 47.92±3.11             & 56.78±5.11             & 86.12±2.91             & 87.84±1.10             & 88.72±1.63             \\
        TCACNet~\cite{R42}        & \scriptsize\textcolor{gray}{[IPM'22]}          & 81.08±4.77                     & 60.25±4.05             & 89.06±2.25             & 78.30±3.68             & 50.17±3.71             & 55.21±3.89             & 87.15±1.87             & 87.85±4.13             & 87.68±3.58             \\
        EEGNet~\cite{R1}          & \scriptsize\textcolor{gray}{[J Neural Eng'18]} & \underline{90.63±1.64}         & 58.86±2.92             & \underline{95.31±1.32} & 81.24±2.31             & 60.77±5.17             & \bf{63.88±5.46}        & 93.06±1.05             & \bf{95.14±1.44}        & \underline{92.19±1.73} \\
        \midrule
        \bf{D-FaST}               & \scriptsize\textcolor{gray}{[\bf{Ours}]}       & \bf{90.98±3.27}                & \underline{67.36±3.08} & \bf{95.84±1.42}        & \bf{83.86±2.40}        & \underline{64.58±2.28} & \underline{63.71±3.03} & \underline{94.10±3.01} & \underline{93.24±3.78} & \bf{94.09±1.68}        \\
        \bottomrule
    \end{tabular}
    \label{table_within_2A}
\end{table*}

\begin{table*}[htbp]
    \centering
    \setlength\tabcolsep{2pt}
    \caption{Performance of within subject experiment with different models on BCIC IV-2B.}
    \begin{tabular}{ccccccccccc}
        \toprule
        \multirow{2}{*}{Model}    & \multirow{2}{*}{Venue}                         & \multicolumn{9}{c}{BCIC IV-2B}                                                                                                                                                                                                         \\
        \cline{3-11}
        \multicolumn{1}{c}{}      & \multicolumn{1}{c}{}                           & B01                            & B02                    & B03                    & B04                    & B05                    & B06                    & B07                    & B08                    & B09                    \\
        \midrule
        RACNN~\cite{R43}          & \scriptsize\textcolor{gray}{[IJCAI'21]}        & 66.00±2.76                     & 52.86±1.26             & 56.38±1.96             & 91.63±7.41             & 56.56±4.60             & 61.25±10.7             & 74.94±4.92             & 88.25±7.64             & 68.19±10.1             \\
        TACNet~\cite{R34}         & \scriptsize\textcolor{gray}{[UbiComp'21]}      & 75.75±0.77                     & \underline{59.50±0.60} & 56.69±0.58             & 97.69±0.98             & 89.50±0.84             & 87.94±0.57             & \bf{84.88±0.65}        & \bf{93.31±0.47}        & 90.00±0.91             \\
        DeepConvNet~\cite{R30}    & \scriptsize\textcolor{gray}{[HBM'17]}          & 73.94±1.98                     & \bf{61.71±1.17}        & \bf{66.31±0.79}        & 97.44±0.26             & 90.94±0.70             & 79.00±1.83             & 81.19±1.28             & 92.00±1.32             & 81.50±0.95             \\
        ShallowConvNet~\cite{R30} & \scriptsize\textcolor{gray}{[HBM'17]}          & 73.84±1.24                     & 56.36±1.50             & 56.25±0.82             & 96.84±0.40             & 87.69±0.52             & 82.00±1.05             & 82.56±0.60             & 90.56±0.64             & 85.13±0.84             \\
        LMDA-Net~\cite{R4}        & \scriptsize\textcolor{gray}{[NeuroImage'23]}   & 74.56±0.90                     & 59.50±2.48             & 61.50±2.14             & 97.81±0.38             & 90.44±1.71             & 84.69±1.94             & 83.50±0.97             & 92.25±1.30             & 86.81±0.81             \\
        EEG-ChannelNet~\cite{R48} & \scriptsize\textcolor{gray}{[TPAMI'21]}        & 63.75±0.65                     & 56.07±0.58             & 53.81±0.55             & 96.81±0.87             & 74.00±0.75             & 73.25±4.53             & 75.00±1.25             & 91.81±0.84             & 85.00±1.80             \\
        SBLEST~\cite{R41}         & \scriptsize\textcolor{gray}{[TPAMI'23]}        & 72.12±0.73                     & 55.93±2.19             & 54.36±1.49             & 92.96±0.34             & 87.96±0.66             & 82.08±0.99             & 76.05±0.70             & 89.52±0.56             & 77.12±0.95             \\
        TCACNet~\cite{R42}        & \scriptsize\textcolor{gray}{[IPM'22]}          & 76.50±1.16                     & 57.38±0.54             & 56.81±1.28             & 97.63±0.42             & 88.56±1.05             & \underline{87.25±0.84} & 83.38±1.39             & \underline{93.06±0.41} & 90.31±1.80             \\
        EEGNet~\cite{R1}          & \scriptsize\textcolor{gray}{[J Neural Eng'18]} & \underline{75.94±0.38}         & 57.50±1.36             & \underline{62.44±1.35} & \bf{98.31±0.17}        & \underline{92.86±0.75} & 87.13±0.71             & \underline{84.38±0.49} & 92.31±0.47             & \underline{92.56±0.56} \\
        \midrule
        \bf{D-FaST}               & \scriptsize\textcolor{gray}{[\bf{Ours}]}       & \bf{78.94±1.18}                & 59.43±2.31             & 60.94±1.96             & \underline{97.94±0.17} & \bf{93.56±0.93}        & \bf{88.50±1.16}        & 83.69±0.84             & 92.44±0.46             & \bf{92.94±0.52}        \\
        \bottomrule
    \end{tabular}
    \label{table_within_2B}
\end{table*}
Numerous hyperparameters are incorporated in D-FaST. To fully harness the potential of D-FaST and identify general rules, we conducted a thorough exploration of the model by varying the collocation hyperparameter settings.

\subsubsection{Number of Features in the Target Frequency Domain}

To prominently showcase the advantages of D-FaST in frequency domain feature extraction, we compared the accuracy of baseline models~\cite{R4, R30, R1} using the same number of frequency domain features on the MNRED dataset. As illustrated in Fig.~\ref{fig3}(a), regardless of the number of features employed in the frequency domain, D-FaST consistently outperforms the other models. It is worth noting that while all models demonstrate performance improvement with an increase in the number of frequency domain features, ShallowConNet and EEGNet approach saturation, whereas D-FaST still exhibits significant room for improvement. Taking into consideration the number of parameters, as well as the time and space complexity associated with training when augmenting the number of features in the frequency domain, we limited our exploration to a maximum of 128 features. For larger datasets and more complex tasks involving decoding brain cognitive signals, further increasing the number of features in the frequency domain can be explored.

\subsubsection{Spatial Sparsity Coefficient and Number of Subspace Nodes}

The spatial sparsity coefficient and number of subspace nodes control the size and range of the dynamic brain connection graph from both the source and target nodes. For example, a coefficient of 1 indicates that the target node's field of view in each subspace is equivalent to that of all source nodes. With only one subspace node, all source node features are compressed into a single target node. In our evaluation on the MNRED dataset, shown in Fig.~\ref{fig3}(b), we tested the model's performance with 7 subspace nodes and 4 spatial sparsity settings. Despite the dataset having only 30 brain leads, we also experimented with 32 and 64 subspace nodes, aligning with the concept of virtual brain regions mentioned in Section~\ref{virtual_regions_of_interest}. Overall, the model's performance initially improves and then declines as the number of subspace nodes increases, reaching maximum average performance at 16 nodes. Moreover, higher accuracy is observed with a small number of subspace nodes (e.g., $N'=1$) and a larger spatial sparsity coefficient, while a smaller sparse coefficient (e.g., $\tau=0.4$) yields greater accuracy in a larger subspace with more nodes (e.g., $N'=64$). The best performance is achieved when these two factors are balanced (e.g., $N'=8$ \& $\tau=0.6$ or $N'=16$ \& $\tau=0.8$).

\subsubsection{Local Temporal Sliding Window}

We examined the performance differences of the model across various datasets at different window size settings, as illustrated in Fig.~\ref{fig3}(c). With a window size of 128, the pooling layer renders the window almost global, implying no window is used. Consequently, the model's sensitivity to this hyperparameter is relatively smaller than that of other hyperparameters. Generally, the model performs optimally at a window size of 64 and exhibits saturation and a decline at larger windows. These finding suggests that utilizing a local temporal sliding window assists the model in efficiently exploring a broader range of attention.
\begin{figure*}[h]
    \centering
    \subfloat[]{\includegraphics[width=0.49\linewidth]{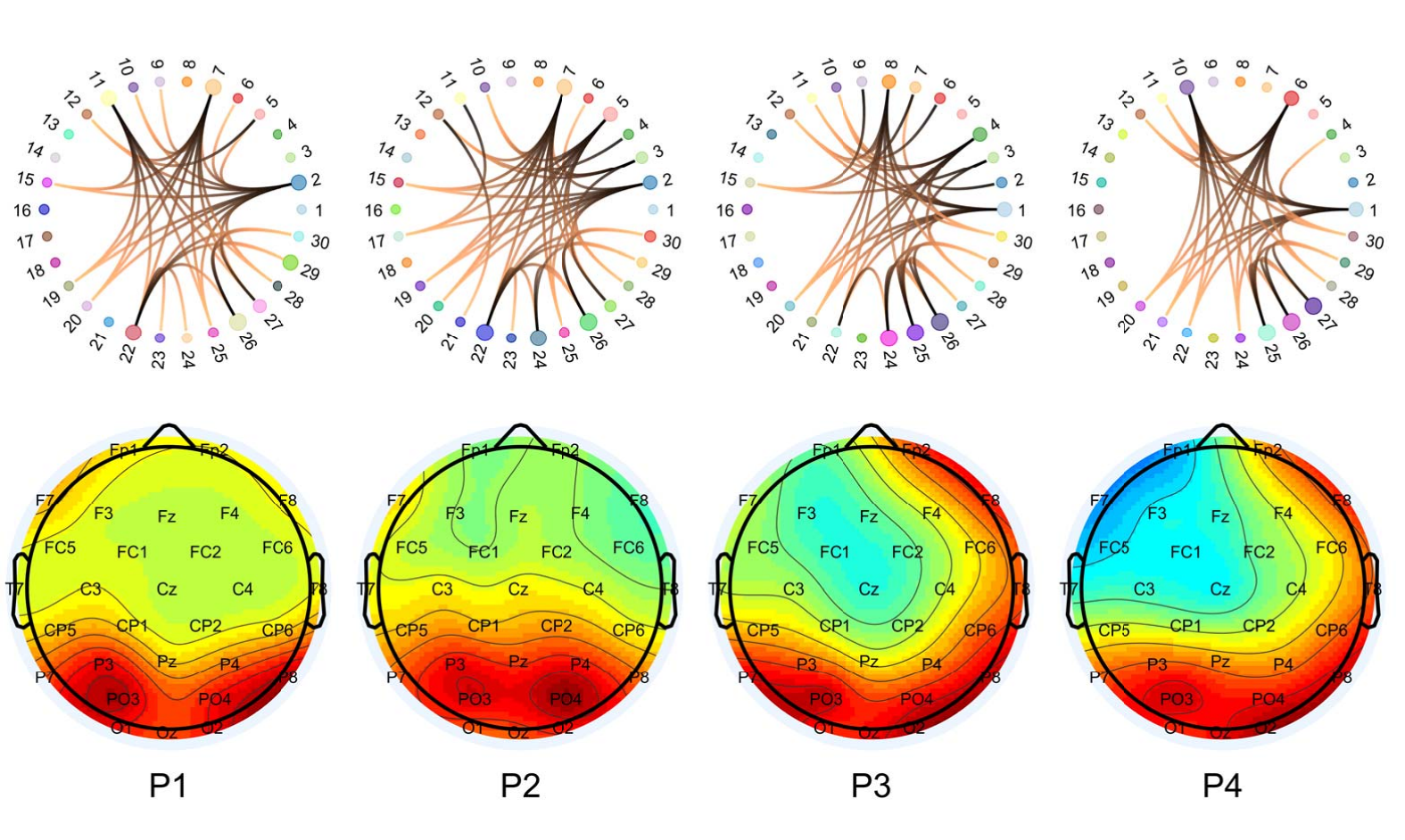}}
    \quad
    \subfloat[]{\includegraphics[width=0.49\linewidth]{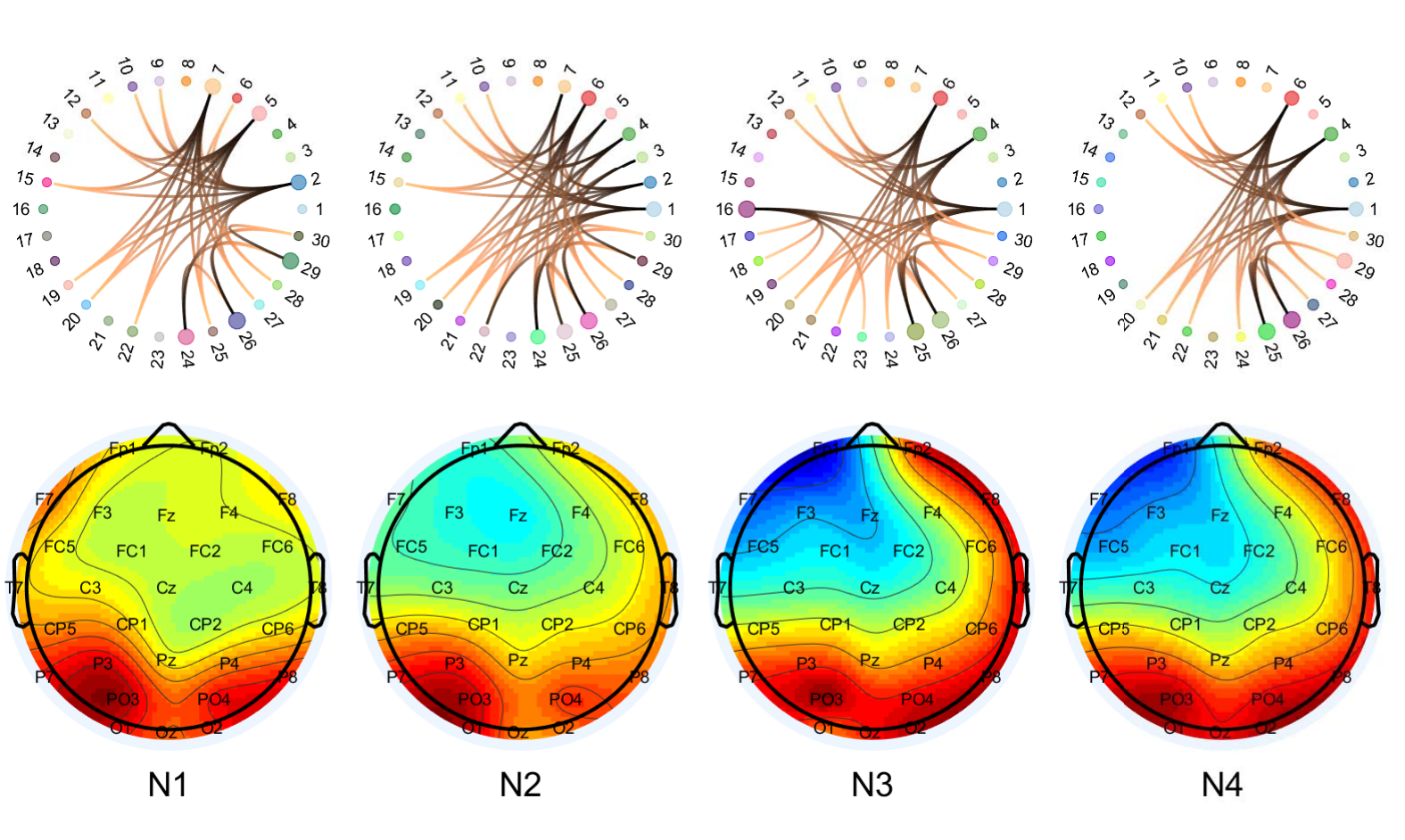}}
    \caption{Dynamic connectogram of negative and positive samples.  (a): Positive examples. (b): Negative examples. It is a description of a cognitive process of the brain consisting of four stages from left to right. The four circular diagrams at the top correspond to the dynamic brain connectivity of each stage, with darker colors indicating stronger weights of the corresponding connection edges. The four heat maps at the bottom correspond to the energy level of brain regions in each stage, with higher heat indicating higher energy level.}
    \label{fig_attention_dca}
\end{figure*}

\begin{figure}[htbp]
    \centering
    \includegraphics[width=1\linewidth]{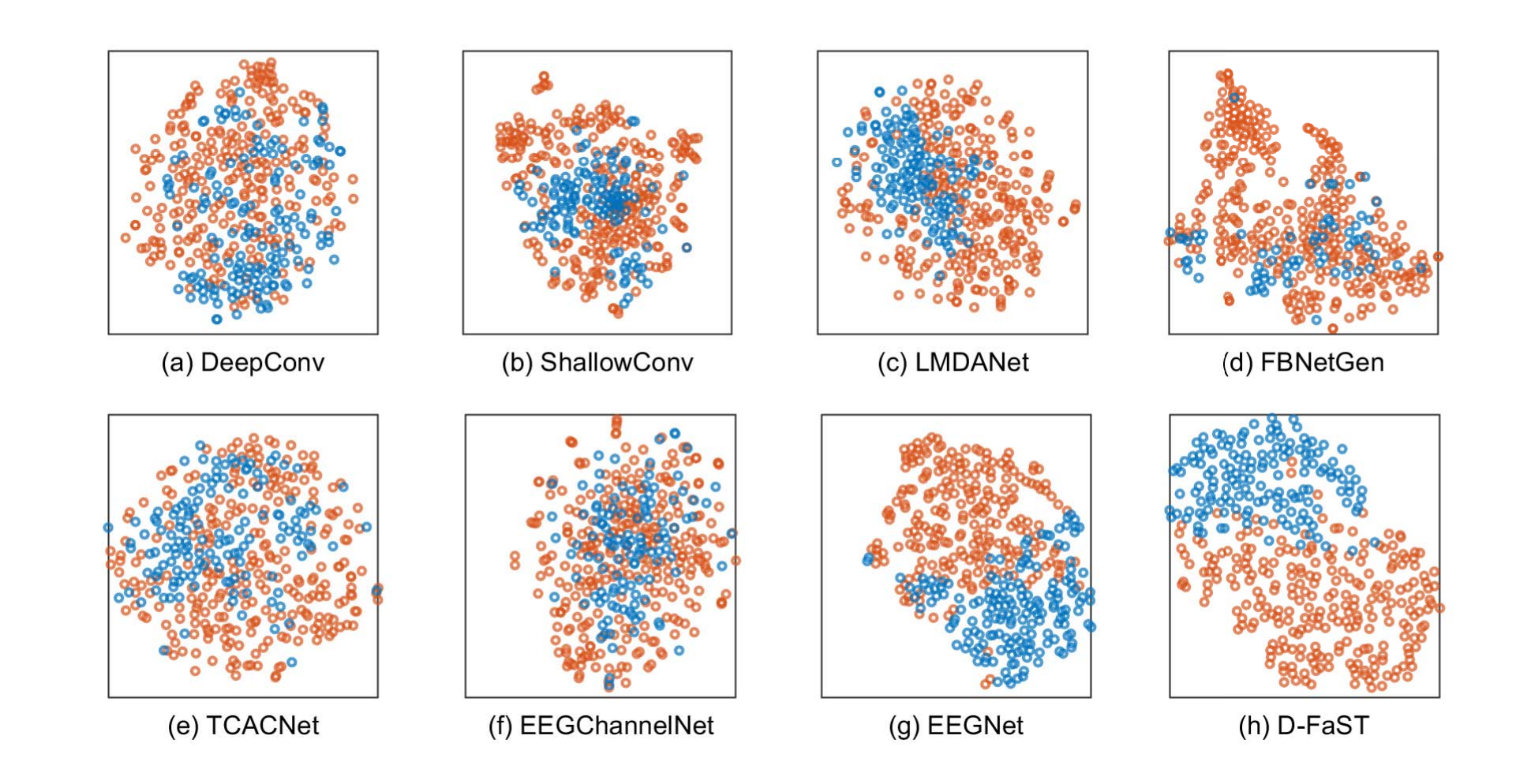}
    \caption{2D projections of the embeddings of different models on MNRED by using t-SNE.}
    \label{fig_t_sne_mnred}
\end{figure}

\subsection{Visualization Analysis of Model Behavior (Q5)}

\subsubsection{Multi-view Attention}
\label{vis_mva}
To provide a more intuitive illustration of the model behavior of multi-view attention, we visualize $\mathcal{A}_F$ in the MVA. Specifically, we input a batch of 128 randomly sampled test data into the trained model and obtain $\mathcal{A}_F$. We then average it from the MVA module and further average the attention weights from the same group of convolutional cores, as shown in Fig.~\ref{fig3}(d). For instance, "49-43"(highlighted in Fig.~\ref{fig3}(d)) refers to frequency domain features of two channels obtained after passing brain cognitive signal data through a two-dimensional convolution layer with a convolution kernel of (1,49), and another two-dimensional convolution layer with a convolution kernel of (1,43). After averaging the attention weights corresponding to these four channels, the weight is 0.5469. We find that using the "49-43" convolution combination, i.e. "$f/4$-$f/4$", yields the maximum weight, followed by "91-1"(the last combination in Fig.~\ref{fig3}(d)), i.e. "$f/2$-1". This suggests that $f/2$ (used in~\cite{R1}) may not be the best choice for the perspective of feature extraction in the frequency domain. The MVA captures the optimal configuration from multiple perspectives in a learnable way. In fact, the weights obtained by these combinations are not significantly different from each other, indicating that the model obtains valuable information from various fields of view.

\subsubsection{Dynamic Connectogram}

In order to provide a more intuitive illustration of the model behavior of dynamic brain connectomogram attention, we visualize $\mathcal{A}_S$ in the DCA. Fig.~\ref{fig_attention_dca} visualizes two sets of brain cognitive models generated by subject 6 (randomly selected) when negative and positive data are observed. The main hyperparameters are set as: $\tau =0.1,N'=30,h=4$(all other parameters are set the same as 4.3). During the initial stage, we note minimal variance between positive and negative brain topographic maps, and the associated brain connection maps are similar. This indicates that subjects, having just been exposed to visual stimuli, hadn't yet distinguished between stimulus categories. In the second stage, these differences start to incrementally increase; by the third stage, notable disparities emerge in both the brain topography and connection maps. We infer that at this juncture, subjects have processed the stimulus and formed subjective judgments. In the fourth stage, although the differences slightly diminish, they continue to persist. This persistence could be due to subjects' uncertainty concerning their judgment after making an initial categorization of the stimulus, leading to continued variation in the brain connectivity map. In summary, the dynamic brain connection map and brain topography map maintain a high level of consistency throughout the cognitive process. This suggests that DCA can dynamically depict different stages of the brain's cognitive process and differentiate between distinct cognitive behaviors.

\section{Conclusion and Analysis}
\label{conclution_analysis}
In this article, we introduce a new CLP dataset called MNRED, featuring a novel paradigm that addresses common issues in brain cognitive signal decoding tasks. We also propose a brain cognitive signal decoder named D-FaST. By innovating the coding mechanisms for frequency domain information, spatial information, and temporal information, as well as designing a decoupled structure for EEG signal processing that better captures the characteristics of relationships between different domains of information, we have significantly enhanced the analysis of EEG signal data.
Through experiments conducted on MNRED, ZuCo, and two classic datasets, BCIC IV-2A and BCIC IV-2B, we have verified the superior performance of our model, achieving state-of-the-art results.

\section*{Acknowledgments}

The work is supported by the Research Project (BHQ090003000X03).

\bibliographystyle{IEEEtran}
\bibliography{references}

\begin{IEEEbiography}[{\includegraphics[width=1in,height=1.25in,clip,keepaspectratio]{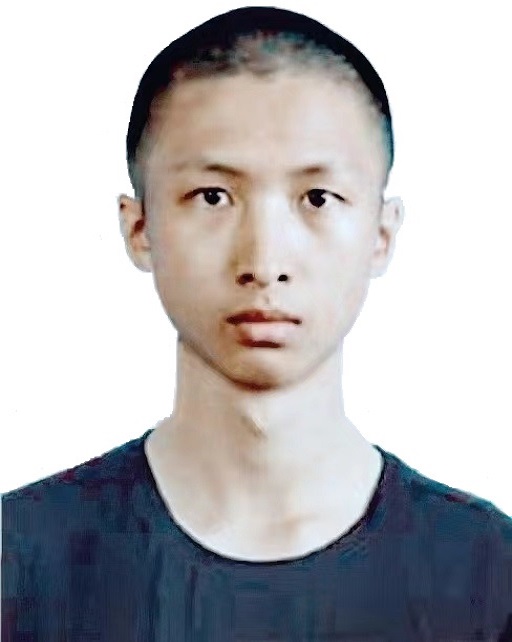}}]{WeiGuo Chen}
received the BE degree in Computer Science and Technology from National University of Defense Technology, China, in 2022, where he is currently pursuing the master's degree. His research interests include cognitive intelligence, multimodal learning in brain-computer interface, time series analysis and natural language processing.
\end{IEEEbiography}

\begin{IEEEbiography}[{\includegraphics[width=1in,height=1.25in,clip,keepaspectratio]{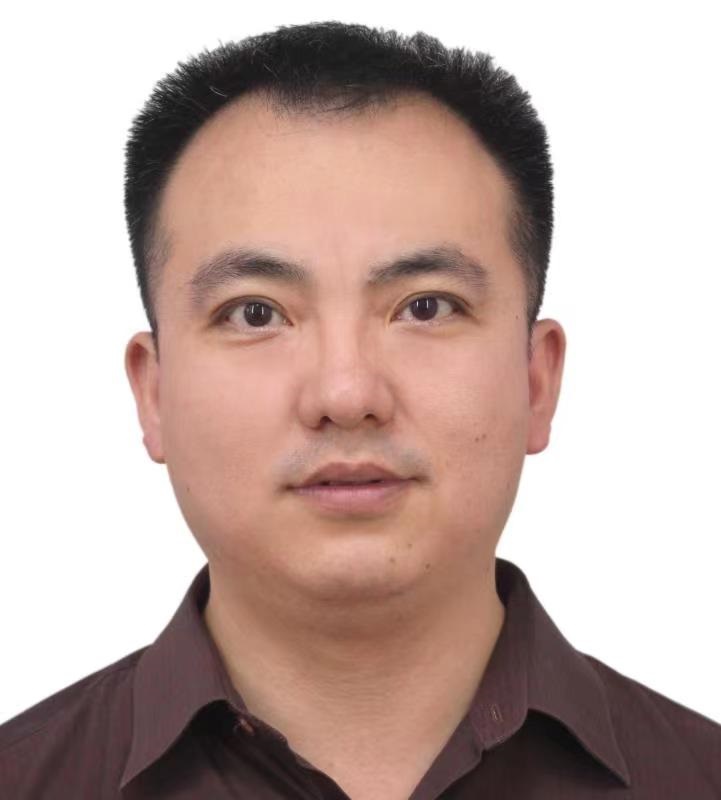}}]{Changjian Wang}
  received his Ph.D. degree in computer science from the School of Computer, National University of Defense Technology. He is currently a Professor of the National University of Defense Technology, Changsha, China.  His current research interests include artificial intelligence, big data and natural language processing.
\end{IEEEbiography}

\begin{IEEEbiography}[{\includegraphics[width=1in,height=1.25in,clip,keepaspectratio]{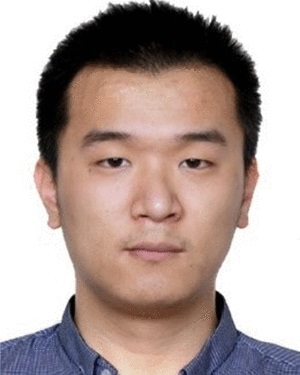}}]{Kele Xu}
 (Member, IEEE) received the doctorate degree in informatique, les télécommunications et l'électronique from Paris VI University, Paris, France, in 2017. He is currently an Associate Professor with the School of Computer Science, National University of Defense Technology, Changsha, China. His research interests include audio signal processing, machine learning, and intelligent software systems.
\end{IEEEbiography}

\begin{IEEEbiography}[{\includegraphics[width=1in,height=1.25in,clip,keepaspectratio]{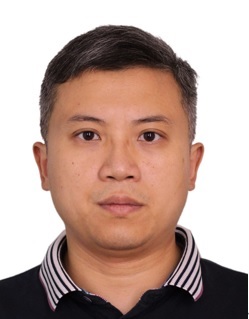}}]{Yuan Yuan}
received the PhD degree at National University of Defense Technology, China. He is now an associate professor at National University of Defense Technology, China. His research interests focus on supercomputer systems, AIOPs and system monitoring and diagnosis.
\end{IEEEbiography}

\begin{IEEEbiography}[{\includegraphics[width=1in,height=1.25in,clip,keepaspectratio]{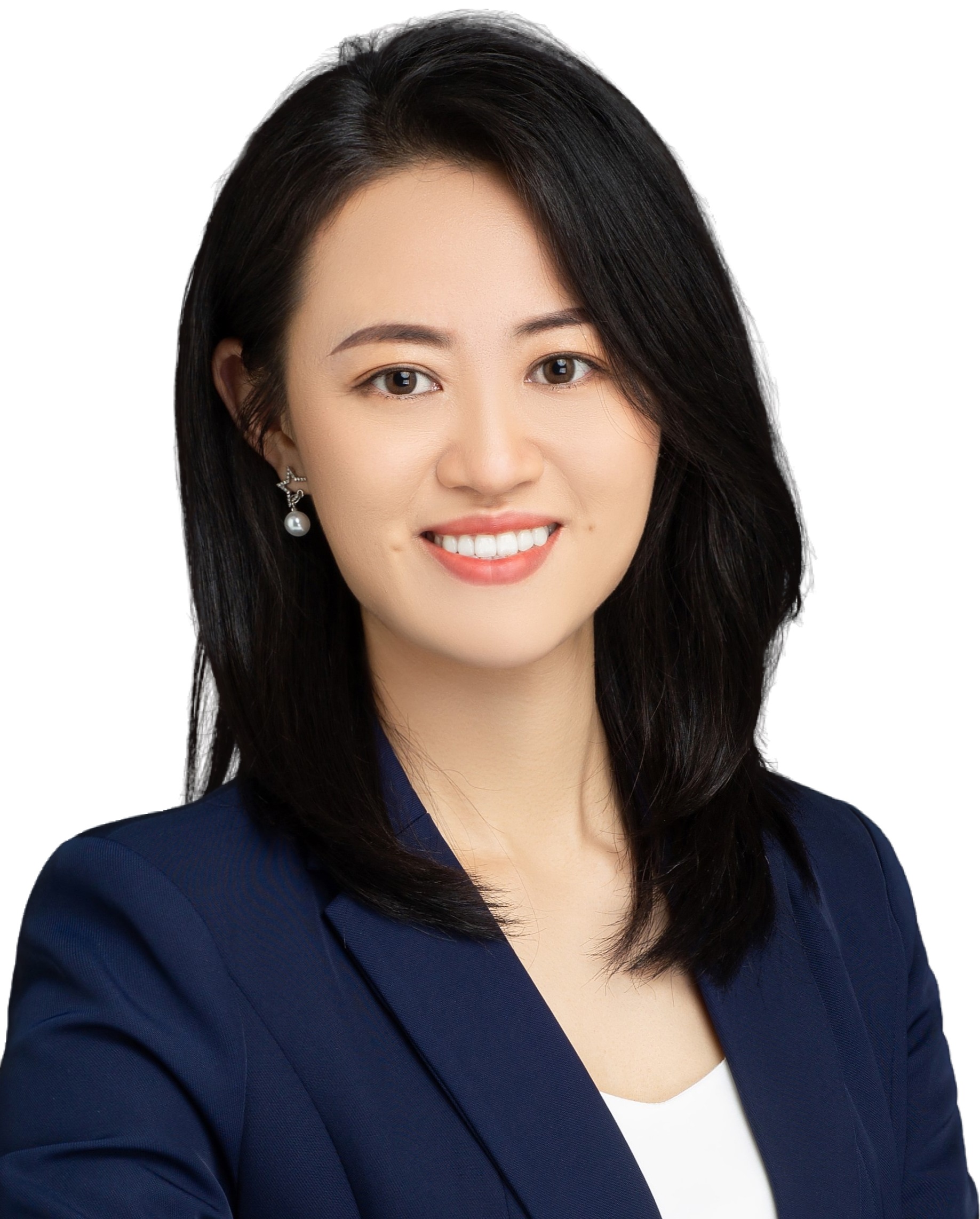}}]{Yanru Bai}
received the B.Eng. and M.Eng. degrees in biomedical engineering from Tianjin University, China, and the Ph.D. degree in biomedical engineering from Nanyang Technological University, Singapore. She is currently an associate professor with Academy of Medical Engineering and Translational Medicine, Tianjin University. Her research interests include brain-computer interface, biomedical signal processing, computational neuroscience, and neural engineering and rehabilitation.
\end{IEEEbiography}

\begin{IEEEbiography}[{\includegraphics[width=1in,height=1.25in,clip,keepaspectratio]{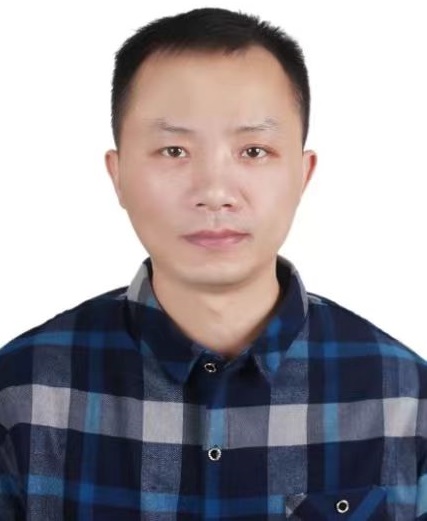}}]{Dongsong Zhang}
received the PhD degree in Computer Science and Technology from National University of Defense Technology, China, in 2012. He currently is an associate professor in School of Big Data andArtificial Intelligence at Xinyang College. His research interests are in the area of soft computing (Neural Network, Fuzzy Logic), Real-Time Systems. 
\end{IEEEbiography}

\appendix

\subsection{Model Details}
The algorithm utilizes multi-view convolution and multi-view attention, which correspond to Eq.\ref{mva_1} and Eq.~\ref{mva_2}, respectively. To ensure that the output and input of a multi-view convolution share the same time dimension, we apply zero padding. Furthermore, we introduce a GELU activation function between the two layers.

These convolution kernels in $\text{InceptionBlock}$ corresponding to Eq.~\ref{dca_1} and Eq.~\ref{dca_2} differ from those in MVA. Although both methods employ convolution kernels of varying sizes to extract rich features, DCA employs fewer kernels compared to MVA. Specifically, the convolution kernels in DCA are small convolutions:$\left[ \left( 1,1 \right) ,\ \left( 1,2 \right) ,\ \left( 1,3 \right) \right] $.

To control the size of model parameters and enhance operational efficiency, we set the number of groups of convolutions in Eq.~\ref{lsta_1} to $E = k \times N'$ to obtai $Q_T,K_T,V_T$, whose nature are relative to DWConv2d. Without loss of generality, we still describe them as CNN. Actually, we can modify its representational power by assigning different values to $E$ and adjusting the dimensions of $Q_T,K_T,V_T$. Additionally, we can enhance the parallelism of matrix multiplication in the algorithm by utilizing multi-head attention. It is important to note that in order to prevent gradient vanishing, we have implemented a residual-like dense structure in LSTA. DCA does not employ residual connections since it already runs in parallel with MVA.

\subsection{Parameter Setting of Baseline Model}
\label{hyperparameter_baseline}

TABLE~\ref{table_hyperparameter_baseline} shows specific hyperparameter settings of comparative models~\cite{R1, R3, R11, R30, R24, R4} reproduced and tuned in this paper. Other parameter settings not mentioned are kept consistent with the original literature. To fairly compare model performance, all models use the same optimizer, learning rate and schedule, minibatch size and number of iterations, and weight decay absorption.

\subsubsection{Number of Model Parameters}

TABLE~\ref{table_parameter_scales} shows the number of parameters for the different models. It can be seen that the number of parameters of the D-FaST model on the MNRED dataset is much larger than that of other models. Nevertheless, the model does not overfit due to the large number of parameters, which indicates the robustness of D-FaST to a certain extent.

\subsubsection{Stratified Sampling \& Cross Validation}

Stratified sampling in training-validation set division often leads to significant disparities in experimental outcomes. Previous studies~\cite{R1,R30} typically separated data into predefined sets, leading to inefficient use of limited brain signal data and potentially skewed model evaluations. Prior attempts~\cite{R4} to reclassify datasets failed to address imbalanced data quality and were discarded. Our method uses stratified sampling techniques~\cite{R11} to balance data category distribution within each cross-validation fold. In cross-subject analyses, it ensures equal data proportions from different subjects across partitions. For within-subject variations, it maintains data distribution equilibrium from different time periods within a dataset. We've also included fixed random seeds in random dataset partitioning to ensure experiment fairness and reproducibility.

\subsection{Visualization detail}
\label{visualization_detail}
Using Fig.~\ref{fig_attention_dca} as an example, the steps for visualization are as follows:
\begin{itemize}
    \item{we used the training sets of all subjects to train the model under the conditions $\tau =0.6,N'=30,h=4$ (other parameter Settings are consistent with Section~\ref{experiment_MNRED});}
    \item{EEG data $\mathcal{X}^{\star}$ where $N=30,T=440$ is obtained by superimposing and averaging the cognitive signals of all negative cases in the pre-treated No.6 (randomly selected) subjects in the test set;}
    \item{On the one hand, we evenly divide $\mathcal{X}^{\star}$ into $h=4$ stages, and then obtain the dynamic energy graph of size $4\times 30\times 110$ and plot it as a brain topographic map, namely the four graphs at the bottom of Fig.~\ref{fig_attention_dca}; On the other hand, we input the model $\mathcal{X}^{\star}$ and extract the set of dynamic connection matrices $\mathcal{A}_S=\{\mathcal{A}_{S}^{1},\mathcal{A}_{S}^{2},\mathcal{A}_{S}^{3},\mathcal{A}_{S}^{4}\}$ from the DCA module, respectively select the channels with the largest multi-view attention in MVA (as described in Section~\ref{vis_mva}), obtain the dynamic brain connection graph, and draw the directed graph respectively, namely the four graphs above Fig.~\ref{fig_attention_dca}. In order to highlight the important parts of the brain connectivity map, we set the sparsity $\tau$ to 0.1 during the test, that is, each node has a maximum $0.1\times 30=3$ of exits.}
\end{itemize}

\subsection{More Results}
\label{more_results}
Although leave-one-subject-out cross-validation is common in CSD, K-fold cross-validation occupies a very important place in classical machine learning. We carried out 5-fold cross-validation on MNRED, ZuCo, BCIC IV-2A and BCIC IV-2B as additional evidence. Results shows in TABLE~\ref{table_MNRED_5f}, TABLE~\ref{table_ZuCo_5f} and TABLE~\ref{table_cross_5f}.

\begin{table*}[htbp]
    \centering
    \setlength\tabcolsep{2.5pt}
    \caption{Parameter scales for different models.}
    \begin{tabular}{c|cccc}
        \toprule
        Model                        & MNRED   & ZuCo   & BCIC IV-2A & BCIC IV-2B \\
        \midrule
        BNT~\cite{R11}               & 232K    & 1,340K & 161K       & 27K        \\
        BrainNetCNN~\cite{R3}        & 172K    & 503K   & 136K       & 51K        \\
        DeepConvNet~\cite{R30}       & 61K     & -      & 73K        & 46k        \\
        ShallowConvNet~\cite{R30}    & 51K     & -      & 41K        & 8K         \\
        FBNetGen~\cite{R24}          & 96K     & 308K   & 80K        & 38K        \\
        LMDA-Net~\cite{R4}           & 9K      & -      & 6K         & 4K         \\
        EEGNet~\cite{R1}             & 726K    & -      & 36K        & 35K        \\
        TACNet~\cite{R34}            & 115K    & -      & 90K        & 28K        \\
        RACNN~\cite{R43}             & 12,180K & -      & 15,227K    & 14,200K    \\
        EEG-ChannelNet~\cite{R48}    & 2,135K  & -      & 2,135K     & 1,204K     \\
        TCACNet~\cite{R42}           & 115K    & -      & 90K        & 28K        \\
        Graph Transformer~\cite{R12} & -       & 739K   & -          & -          \\
        \bf{D-FaST(ours)}            & 4,302K  & 151K   & 672K       & 11K        \\
        \bottomrule
    \end{tabular}
    \label{table_parameter_scales}
\end{table*}
\begin{table*}[htbp]
    \centering
    \setlength\tabcolsep{2pt}
    \caption{Baseline Model hyperparameter settings.}
    \begin{tabular}{c|cccccccccccccccccccc}
        \midrule
        Model                        & \multicolumn{20}{c}{Hyper-Parameter}                                                                                                                                                   \\
        \midrule
        \multirow{2}{*}{EEGNet}      & \multicolumn{4}{c}{Num kernels}      & \multicolumn{4}{c}{P1}              & \multicolumn{4}{c}{D}              & \multicolumn{4}{c}{P2}          & \multicolumn{4}{c}{Dropout}        \\
                                     & \multicolumn{4}{c}{16}               & \multicolumn{4}{c}{4}               & \multicolumn{4}{c}{2}              & \multicolumn{4}{c}{8}           & \multicolumn{4}{c}{0.5}            \\
        \midrule
        \multirow{2}{*}{LMDA-Net}    & \multicolumn{4}{c}{Channel depth1}   & \multicolumn{4}{c}{Ave depth}       & \multicolumn{4}{c}{Depth}          & \multicolumn{4}{c}{K}           & \multicolumn{4}{c}{Channel depth2} \\
                                     & \multicolumn{4}{c}{product}          & \multicolumn{4}{c}{5}               & \multicolumn{4}{c}{9}              & \multicolumn{4}{c}{7}           & \multicolumn{4}{c}{16}             \\
        \midrule
        \multirow{2}{*}{BNT}         & \multicolumn{4}{c}{Pooling}          & \multicolumn{4}{c}{Freeze center}   & \multicolumn{4}{c}{Sizes}          & \multicolumn{4}{c}{Dim}         & \multicolumn{4}{c}{Orthogonal}     \\
                                     & \multicolumn{4}{c}{(False, True)}    & \multicolumn{4}{c}{True}            & \multicolumn{4}{c}{(N, N/2)}       & \multicolumn{4}{c}{1024}        & \multicolumn{4}{c}{True}           \\
        \midrule
        \multirow{2}{*}{FBNetGen}    & \multicolumn{5}{c}{Extractor type}   & \multicolumn{5}{c}{Graphgeneration} & \multicolumn{5}{c}{Embedding size} & \multicolumn{5}{c}{Window size}                                      \\
                                     & \multicolumn{5}{c}{cnn}              & \multicolumn{5}{c}{product}         & \multicolumn{5}{c}{16}             & \multicolumn{5}{c}{50}                                               \\
        \midrule
        \multirow{2}{*}{BrainNetCNN} & \multicolumn{5}{c}{E2E1}             & \multicolumn{5}{c}{E2E2}            & \multicolumn{5}{c}{E2N}            & \multicolumn{5}{c}{N2G}                                              \\
                                     & \multicolumn{5}{c}{(1, 32)}          & \multicolumn{5}{c}{(32, 64)}        & \multicolumn{5}{c}{(64, 1)}        & \multicolumn{5}{c}{(1, 256)}                                         \\
        \midrule
                                     & \multicolumn{10}{c}{Num kernels}     & \multicolumn{10}{c}{Dropout}                                                                                                                    \\
        DeepConvNet                  & \multicolumn{10}{c}{25}              & \multicolumn{10}{c}{0.5}                                                                                                                        \\
        ShallowConvNe                & \multicolumn{10}{c}{40}              & \multicolumn{10}{c}{0.5}                                                                                                                        \\
        \bottomrule
    \end{tabular}
    \label{table_hyperparameter_baseline}
\end{table*}

\begin{table*}[htbp]
    \centering
    \caption{Compare experimental results under cross-subject (5-fold corss-validation) experimental settings on MNRED.}
    \begin{tabular}{cccccccc}
        \toprule
        \multirow{2}{*}{Model}                        & \multirow{2}{*}{Venue}                         & \multicolumn{1}{c}{\multirow{2}{*}{Type}}      & \multicolumn{4}{c}{MNRED}                                                                                                \\
        \cline{4-7}
        \multicolumn{1}{c}{}                          & \multicolumn{1}{c}{}                           & \multicolumn{1}{c}{}                           & \makecell[c]{Accuracy (\%)} & \makecell[c]{AUROC(\%)} & \makecell[c]{Sensitivity  (\%)} & \makecell[c]{Specificity (\%)} \\
        \midrule
        \multicolumn{1}{c}{BrainNetCNN~\cite{R3}}     & \scriptsize\textcolor{gray}{[NeuroImage'17]}   & \multicolumn{1}{c}{\multirow{4}{*}{1D}}        & 73.88±2.48                  & 72.15±1.44              & 37.30±8.03                      & 81.24±1.52                     \\
        \multicolumn{1}{c}{BNT~\cite{R11}}            & \scriptsize\textcolor{gray}{[NeurIPS'22]}      & \multicolumn{1}{c}{}                           & 74.11±2.37                  & 72.30±2.24              & 37.61±10.34                     & 80.15±2.26                     \\
        \multicolumn{1}{c}{TACNet~\cite{R34}}         & \scriptsize\textcolor{gray}{[UbiComp'21]}      & \multicolumn{1}{c}{}                           & 79.66±0.73                  & 81.68±2.69              & 62.86±7.03                      & 86.79±2.23                     \\
        \multicolumn{1}{c}{RACNN~\cite{R43}}          & \scriptsize\textcolor{gray}{[IJCAI'21]}        & \multicolumn{1}{c}{}                           & 79.33±0.80                  & 80.57±0.82              & 61.57±0.65                      & 86.94±0.89                     \\
        \midrule
        \multicolumn{1}{c}{DeepConvNet~\cite{R30}}    & \scriptsize\textcolor{gray}{[HBM'17]}          & \multicolumn{1}{c}{\multirow{7}{*}{2D-Serial}} & 77.45±0.64                  & 82.90±1.13              & 72.03±3.55                      & 79.88±2.58                     \\
        \multicolumn{1}{c}{ShallowConvNet~\cite{R30}} & \scriptsize\textcolor{gray}{[HBM'17]}          & \multicolumn{1}{c}{}                           & 80.31±1.04                  & 84.70±1.17              & 67.33±0.40                      & 85.92±1.77                     \\
        \multicolumn{1}{c}{FBNetGen~\cite{R24}}       & \scriptsize\textcolor{gray}{[MIDL'22]}         & \multicolumn{1}{c}{}                           & 78.95±1.80                  & 81.35±1.05              & 54.56±6.47                      & 86.51±1.96                     \\
        \multicolumn{1}{c}{LMDA-Net~\cite{R4}}        & \scriptsize\textcolor{gray}{[NeuroImage'23]}   & \multicolumn{1}{c}{}                           & 78.60±0.72                  & \underline{84.76±0.96}  & \underline{72.73±2.58}          & 81.13±1.52                     \\
        \multicolumn{1}{c}{EEG-ChannelNet~\cite{R48}} & \scriptsize\textcolor{gray}{[TPAMI'21]}        & \multicolumn{1}{c}{}                           & \underline{80.88±0.83}      & 83.98±0.86              & 60.28±0.63                      & \bf{89.83±0.91}                \\
        \multicolumn{1}{c}{TCACNet~\cite{R42}}        & \scriptsize\textcolor{gray}{[IPM'22]}          & \multicolumn{1}{c}{}                           & 80.02±0.96                  & 82.12±1.39              & 63.99±1.49                      & 86.98±1.32                     \\
        \midrule
        \multicolumn{1}{c}{EEGNet~\cite{R1}}          & \scriptsize\textcolor{gray}{[J Neural Eng'18]} & \multicolumn{1}{c}{3D-Serial}                  & 76.22±1.40                  & 83.71±1.92              & \bf{78.82±0.68}                 & 74.87±2.70                     \\
        \midrule
        \multicolumn{1}{c}{\bf{D-FaST}}               & \scriptsize\textcolor{gray}{[\bf{Ours}]}       & \multicolumn{1}{c}{3D-Disentangled}            & \bf{82.96±2.04}             & \bf{87.43±1.85}         & 70.79±5.26                      & \underline{88.17±1.80}         \\
        \bottomrule
    \end{tabular}
    \label{table_MNRED_5f}
\end{table*}

\begin{table*}[htbp]
    \centering
    \setlength\tabcolsep{3pt}
    \caption{Compare experimental results under cross-subject (5-fold corss-validation) experimental settings.}
    \begin{tabular}{cccc}
        \toprule
        \multirow{2}{*}{Model}                           & \multirow{2}{*}{Venue}                       & \multicolumn{2}{c}{ZuCo}                              \\
        \cline{3-4}
        \multicolumn{1}{c}{}                             & \multicolumn{1}{c}{}                         & \makecell[c]{Accuracy (\%)} & \makecell[c]{AUROC(\%)} \\
        \midrule
        \multicolumn{1}{c}{FBNetGen~\cite{R24}}          & \scriptsize\textcolor{gray}{[MIDL'22]}       & 70.30±0.74                  & 90.56±0.92              \\
        \multicolumn{1}{c}{BrainNetCNN~\cite{R3}}        & \scriptsize\textcolor{gray}{[NeuroImage'17]} & 85.49±0.87                  & \underline{93.87±0.9}4  \\
        \multicolumn{1}{c}{Graph Transformer~\cite{R12}} & \scriptsize\textcolor{gray}{[AAAI'21]}       & 87.10±0.88                  & 93.75±0.94              \\
        \multicolumn{1}{c}{BNT~\cite{R11}}               & \scriptsize\textcolor{gray}{[NeurIPS'22]}    & \underline{87.45±0.89}      & \bf{94.05±0.95}         \\
        \midrule
        \multicolumn{1}{c}{\bf{D-FaST}}                  & \scriptsize\textcolor{gray}{[\bf{Ours}]}     & \bf{88.68±0.89}             & 92.77±0.94              \\
        \bottomrule
    \end{tabular}
    \label{table_ZuCo_5f}
\end{table*}

\begin{table*}[htbp]
    \centering
    \setlength\tabcolsep{2.5pt}
    \caption{Compare experimental results under cross-subject (5-fold corss-validation) experimental settings.}
    \begin{tabular}{ccccccccccc}
        \toprule
        \multirow{2}{*}{Model}                        & \multirow{2}{*}{Venue}                         & \multicolumn{1}{c}{\multirow{2}{*}{Type}}      & \multicolumn{2}{c}{BCIC IV-2A} &                           & \multicolumn{4}{c}{BCIC IV-2B}                                                                                                                              \\
        \cline{4-5} \cline{7-10}
        \multicolumn{1}{c}{}                          & \multicolumn{1}{c}{}                           & \multicolumn{1}{c}{}                           & \makecell[c]{Accuracy (\%)}    & \makecell[c]{AUROC  (\%)} &                                & \makecell[cc]{Accuracy (\%)} & \makecell[cc]{AUROC(\%)} & \makecell[c]{Sensitivity  (\%)} & \makecell[c]{Specificity (\%)} \\
        \midrule
        \multicolumn{1}{c}{BrainNetCNN~\cite{R3}}     & \scriptsize\textcolor{gray}{[NeuroImage'17]}   & \multicolumn{1}{c}{\multirow{4}{*}{1D}}        & 54.40±1.10                     & 77.34±0.40                &                                & -                            & -                        & -                               & -                              \\
        \multicolumn{1}{c}{BNT~\cite{R11}}            & \scriptsize\textcolor{gray}{[NeurIPS'22]}      & \multicolumn{1}{c}{}                           & 55.77±0.95                     & 78.77±0.72                &                                & -                            & -                        & -                               & -                              \\
        \multicolumn{1}{c}{TACNet~\cite{R34}}         & \scriptsize\textcolor{gray}{[UbiComp'21]}      & \multicolumn{1}{c}{}                           & 65.99±0.67                     & 85.25±0.58                &                                & 77.04±0.77                   & 85.71±0.86               & 73.72±0.77                      & 80.37±0.84                     \\
        \multicolumn{1}{c}{RACNN~\cite{R43}}          & \scriptsize\textcolor{gray}{[IJCAI'21]}        & \multicolumn{1}{c}{}                           & -                              & -                         &                                & 72.07±0.77                   & 79.84±0.86               & 69.25±0.75                      & 74.89±0.80                     \\
        \midrule
        \multicolumn{1}{c}{DeepConvNet~\cite{R30}}    & \scriptsize\textcolor{gray}{[HBM'17]}          & \multicolumn{1}{c}{\multirow{7}{*}{2D-Serial}} & 71.95±0.50                     & \bf{90.68±0.21}           &                                & 77.60±0.44                   & 86.21±0.64               & 76.24±4.44                      & 78.96±4.50                     \\
        \multicolumn{1}{c}{ShallowConvNet~\cite{R30}} & \scriptsize\textcolor{gray}{[HBM'17]}          & \multicolumn{1}{c}{}                           & \underline{72.38±0.73}         & 90.49±0.42                &                                & 76.66±0.26                   & 85.48±0.08               & 74.72±2.61                      & 78.61±2.65                     \\
        \multicolumn{1}{c}{FBNetGen~\cite{R24}}       & \scriptsize\textcolor{gray}{[MIDL'22]}         & \multicolumn{1}{c}{}                           & 67.26±1.58                     & 87.14±0.61                &                                & 72.05±0.44                   & 80.85±0.87               & 72.68±3.97                      & 71.42±4.62                     \\
        \multicolumn{1}{c}{LMDA-Net~\cite{R4}}        & \scriptsize\textcolor{gray}{[NeuroImage'23]}   & \multicolumn{1}{c}{}                           & 70.22±1.31                     & 88.35±0.63                &                                & 79.68±0.26                   & 88.00±0.37               & \bf{79.52±2.90}                 & 79.85±2.71                     \\
        \multicolumn{1}{c}{EEG-ChannelNet~\cite{R48}} & \scriptsize\textcolor{gray}{[TPAMI'21]}        & \multicolumn{1}{c}{}                           & 61.05±0.63                     & 83.18±8.4                 &                                & 74.98±0.76                   & 83.77±0.85               & 73.04±0.77                      & 76.92±0.813                    \\
        \multicolumn{1}{c}{SBLEST~\cite{R41}}         & \scriptsize\textcolor{gray}{[TPAMI'23]}        & \multicolumn{1}{c}{}                           & -                              & -                         &                                & 72.70±0.24                   & 81.69±0.16               & 67.21±0.31                      & 78.76±0.40                     \\
        \multicolumn{1}{c}{TCACNet~\cite{R42}}        & \scriptsize\textcolor{gray}{[IPM'22]}          & \multicolumn{1}{c}{}                           & 67.50±0.69                     & 85.65±0.87                &                                & 76.62±0.77                   & 85.78±0.86               & 72.08±0.76                      & 81.15±0.84                     \\
        \midrule
        \multicolumn{1}{c}{EEGNet~\cite{R1}}          & \scriptsize\textcolor{gray}{[J Neural Eng'18]} & \multicolumn{1}{c}{3D-Serial}                  & 70.52±1.07                     & 88.82±0.77                &                                & \underline{80.23±0.28}       & \underline{88.91±0.24}   & 79.08±1.24                      & \underline{81.38±1.04}         \\
        \midrule
        \multicolumn{1}{c}{\bf{D-FaST}}               & \scriptsize\textcolor{gray}{[\bf{Ours}]}       & \multicolumn{1}{c}{3D-Disentangled}            & \bf{74.95±0.66}                & \underline{90.56±0.71}    &                                & \bf{80.72±0.31}              & \bf{89.27±0.33}          & \underline{79.48±2.33}          & \bf{81.96±1.92}                \\
        \bottomrule
    \end{tabular}
    \label{table_cross_5f}
\end{table*}
\end{document}